\documentclass[preprint,12pt]{elsarticle}

\usepackage{amsmath,amssymb,amsfonts}
\usepackage{algorithm}
\usepackage{algpseudocode}
\usepackage{graphicx}
\usepackage{textcomp}
\usepackage{xcolor}
\usepackage{hyperref}
\usepackage{booktabs}
\usepackage{multirow}
\usepackage{array}
\usepackage{caption}
\usepackage{subcaption}
\usepackage{float}
\usepackage{makecell}
\usepackage{enumitem}
\usepackage{tabularx}
\usepackage{adjustbox}
\usepackage{rotating}
\usepackage{threeparttable}
\usepackage{stfloats}
\usepackage{placeins}
\usepackage[utf8]{inputenc}
\usepackage{newunicodechar}
\newunicodechar{≥}{\ensuremath{\geq}}

\hypersetup{
    colorlinks=true,
    linkcolor=blue,
    filecolor=magenta,
    urlcolor=cyan,
}

\algrenewcommand{\algorithmicrequire}{\textbf{Input:}}
\algrenewcommand{\algorithmicensure}{\textbf{Output:}}

\makeatletter
\renewcommand{\ALG@beginalgorithmic}{\setcounter{ALG@line}{1}}
\makeatother

\begin{document}

\begin{frontmatter}

\title{Privacy-Preserving Federated Temporal Graph Learning with Digital Twin--Guided Adaptive Deception for Cyber-Resilient IoMT}

\author[label1]{Syed Zeeshan Haider}
\ead{zeeshan.haider.syed1010@gmail.com}

\author[label1]{Anwar Shah\corref{cor1}}
\ead{anwar.shah@nu.edu.pk}

\author[label1]{Muneeb Arif}
\ead{muneebrajpoot9797@gmail.com}

\author[label1]{Hamza Iftikhar}
\ead{hamzaiftikharbhatti@gmail.com}

\author[label1]{Waqas Ali}
\ead{waqas.ali@nu.edu.pk}

\cortext[cor1]{Corresponding author}

\address[label1]{FAST National University of Computer and Emerging Sciences, Islamabad, Pakistan}

\begin{abstract}
The rapid proliferation of Internet of Things (IoT) and Internet of Medical Things (IoMT) devices introduces critical cybersecurity vulnerabilities in healthcare and industrial environments where resource-constrained devices operate under strict latency requirements and stringent data-privacy regulations. This paper presents the Federated Temporal Graph Convolutional Network with Advantage Actor-Critic (Federated TGCN-A2C) framework, a privacy-preserving defense architecture integrating four complementary mechanisms: a PyG-based Temporal GCN using GCNConv layers with global mean pooling and a learned anomaly gate for flow-level threat classification; lightweight LSTM-based Digital Twins trained on normal-class traffic to generate per-device anomaly scores that gate the GCN classifier through a learned sigmoid coupling; a Federated A2C agent selecting among ALLOW, ISOLATE, and HONEYPOT\_REDIRECT actions based on a seven-dimensional state capturing classifier confidence, entropy, anomaly magnitude, and traffic composition; and an intelligent enhanced honeypot layer that converts redirected suspicious traffic into actionable threat intelligence with adaptive capture thresholds. Federated aggregation employs exponential moving average (EMA)-smoothed per-client validation losses as inverse-weighted FedAvg coefficients to stabilise global model updates under non-IID client distributions, with cosine-annealed local learning rates per round. On CICDDoS~2019 and TON-IoT benchmarks, the framework achieves 99.48\% and 99.61\% test accuracy with weighted-F1 of 0.9948 and 0.9961, respectively, converging within 25 and 10 federated rounds, outperforming the Fed-Inforce-Fusion baseline by 0.21 percentage points while covering three additional attack categories. All sixteen classes of CICDDoS~2019 achieve F1~≥$\geq$
≥~0.9237, and all ten classes of TON-IoT achieve F1~≥$\geq$
≥~0.9488, including the severely imbalanced \textit{mitm} category. Post-hoc explainability via SHAP, LIME, Grad-CAM, and counterfactual analysis confirms that classification decisions are grounded in semantically meaningful flow features, supporting the regulatory accountability in clinical deployments. The framework demonstrates that a unified federated architecture combining graph-convolutional temporal modelling, reinforcement-learning-based adaptive response, Digital Twin anomaly gating, and intelligent honeypot deception can achieve state-of-the-art multi-class intrusion detection while preserving data privacy and operational continuity.
\end{abstract}

\begin{keyword}
Internet of Medical Things, Digital Twins, Federated Learning, Graph Convolutional Networks,
Advantage Actor-Critic, Honeypot Defense, Cyber Resilience, Edge Computing,
Intrusion Detection, Non-IID Federated Aggregation, CICDDoS~2019, TON-IoT
\end{keyword}
\end{frontmatter}

\section{Introduction}
\label{sec:introduction}

{T}{he} Internet of Things (IoT) and the Internet of Medical Things (IoMT) represent one of the most consequential paradigm shifts in modern industry and healthcare, bringing together smart sensors, industrial controllers, implantable cardiac monitors, smart insulin pumps, connected infusion systems, continuous vital-sign wearables, and hospital-grade diagnostic equipment into a unified, networked fabric \cite{alcaraz2022,qi2023}. The scope of this transformation is substantial. Market analysts the project that by ~2025, the number of the active IoT and IoMT endpoints worldwide will exceed 5.8~billion, and the IoMT-enabled efficiencies could generate up to the \$300~billion in the healthcare savings \cite{khan2022}. Yet the same connectivity that allows a cardiologist to review a patient's rhythm data from across the city also opens pathways for adversaries, those adversaries may seek to disrupt care, exfiltrate protected health information, or manipulate the device behaviour in ways that directly endanger lives.

The security challenge in IoT and in IoMT is not purely technical; it is architectural. Devices are built under constraints that have historically treated security as secondary to an operational function. Consider typical IoMT endpoints: they operate with processors in the 100--500~MHz range, carry 64--512~MB of working memory, run on battery budgets measured in milliwatts, and must maintain a deterministic real-time response for the life-critical functions such as pacemaker output or insulin bolus delivery \cite{zhang2024iot}. These constraints severely limit the cryptographic and computational load that a device can sustain. At the same time, the data these devices generate is governed by the strict regulatory frameworks, HIPAA in the United States, GDPR in the European Union, and the analogous instruments in most jurisdictions, which impose tight restrictions on where the patient data may be stored, processed, and transmitted \cite{mothukuri2021survey}. The combination of resource scarcity and regulatory burden creates a security environment that standard enterprise intrusion detection technologies cannot address directly.

The threat landscape facing IoT and IoMT deployments has grown substantially in sophistication. Consider false-data injection attacks: they manipulate the sensor readings, causing systems to administer the incorrect dosages or withhold critical interventions. Distributed denial-of-service (DDoS) campaigns flood network segments with amplified traffic, Severing communication between monitoring systems and clinical staff. In Man-in-the-middle interception, adversaries can silently modify the data flows between device and its management pltform. Ransomware targeting the medical infrastructure has emerged as a particularly dangerous vector: the 2020 Duesseldorf University Hospital attack illustrates that the cyber threats to the healthcare infrastructure carry real potential for patient harm \cite{li2023iomt}. Scanning and the password-spraying campaigns continuously probe the hospital networks for exposed management interfaces, as attackers seek footholds for deeper penetration through such interfaces.

When a traditional intrusion detection system is deployed in an IoT context or an IoMT context, it falls short along several dimensions. A signature-based system, though fast, requires a continuous manual rule update and remains fundamentally reactive. Such a system can detect only a known attack pattern and stays blind to a novel variant \cite{nawrocki2023}. A centralised machine learning approach introduces a round-trip latency of 60--120~ms. This latency exceeds a response window of a time-sensitive clinical system \cite{zhang2022fl}, and such an approach also requires that a sensitive patient data travel across a network boundary. This requirement directly conflicts with a data-residency requirement under the HIPAA and the GDPR \cite{singh2023}. Even a cloud-adjacent fog deployment requires a raw data to leave a device, which makes it legally problematic for many healthcare jurisdictions.

A realistic pathway to a security architecture that addresses these constraints simultaneously comes from the convergence of several complementary technologies. \textit{Graph Convolutional Networks} (GCNs) operating on the temporal windows of network flow statistics allow the model to exploit both the feature interactions and time-order structure within each observation window, thereby capturing the relational signatures that distinguish the attack categories \cite{nguyen2023survey}. \textit{Digital Twins} (DTs) create lightweight, continuously synchronised virtual replicas of the physical devices, enabling the real-time behavioural monitoring through prediction-error analysis rather than the raw data inspection \cite{alcaraz2022,khan2022}. \textit{Federated Learning} (FL), the training process is distributed across participating institutions. This allows each site to improve a shared global model without ever transmitting the raw patient records \cite{mcmahan2017communication,mothukuri2021survey}. \textit{Advantage Actor-Critic} (A2C) reinforcement learning equips the defense agents with adaptive response policies that continuously improve as the threat environment evolves \cite{li2024fedrl}. \textit{Honeypot systems} have matured from passive observation traps into active threat intelligence engines that simultaneously waste the attacker's resources and harvest the forensic data on novel techniques \cite{huang2024honeypot}.

A substantial body of recent work has attempted to address IoT and IoMT security through various combinations of these enabling technologies. Popoola et al.\ \cite{popoola2021} demonstrated federated zero-day botnet detection but required 92--100 communication rounds per experiment, making the approach impractical for hospitals where global model updates must be issued within minutes of a new campaign's emergence. Naeem et al.\ \cite{naeem2023} proposed a semi-supervised active learning framework that needed in excess of 100,000 local epochs, a compute budget wholly incompatible with clinical edge hardware. The DTFL-CD system \cite{salim2022} combined Digital Twins with federated learning and achieved F1 scores of 0.98 on CICDDoS~2019, but still required 48~aggregation rounds with CPU utilisation swings between 44\% and 71\%, and offered no adaptive response mechanism beyond raising an alert. The closest prior integrated work, Fed-Inforce-Fusion \cite{khan2022a}, combined federated learning with Q-learning-based reinforcement for IoMT intrusion detection, achieving 99.40\% accuracy on TON-IoT across seven attack classes. Yet Q-learning's tabular value function does not scale to continuous state spaces, the system does not incorporate Digital Twin monitoring to detect behavioural anomalies, and there is no honeypot integration to convert detections into actionable threat intelligence. Most fundamentally, none of these systems address cyber resilience as a first-class design objective: they measure detection accuracy in isolation but do not specify how the system maintains clinical continuity during an active attack or recovers after containment.

To address these compounded limitations, this paper proposes the \textbf{Federated Temporal GCN with A2C and Digital Twin (Federated TGCN-A2C)} framework, a unified, privacy-preserving security architecture designed from the ground up for cyber-resilient IoT and IoMT deployments. The Federated TGCN-A2C integrates four tightly coupled components: lightweight LSTM-based Digital Twins trained on normal-class traffic to produce per-device anomaly scores; a Temporal GCN using PyG's GCNConv layers with global mean pooling whose classifier logits are directly gated by the Digital Twin's reconstruction error through a learned per-class sigmoid coupling; a Federated A2C agent that selects ALLOW, ISOLATE, or HONEYPOT\_REDIRECT actions from a seven-dimensional state encoding classifier confidence, anomaly magnitude, and local traffic composition; and an Enhanced Honeypot System with adaptive capture thresholds that converts suspicious redirected traffic into forensic threat intelligence. Federated training uses EMA-weighted FedAvg to stabilize global model updates under non-IID client distributions, while at the same time, the entire system is governed by a four-level graduated degradation strategy that guarantees clinical operational continuity at every threat level. The framework is evaluated on CICDDoS~2019 and TON-IoT benchmarks, achieving 99.48\% and 99.61\% test accuracy, respectively, converging in 25 and 10 federated rounds with strong per-class performance, including robust detection of the severely underrepresented \textit{mitm} class.

The principal contributions of this work are as follows:
\begin{itemize}[leftmargin=*, noitemsep]
    \item An anomaly-gated Temporal GCN architecture is proposed to combine lightweight Digital Twins with graph-based anomaly detection for improved attack identification.
    \item An EMA-weighted federated aggregation mechanism is introduced to enhance training stability and convergence under non-IID data distributions.
    \item An integrated A2C-based honeypot defense layer is developed to automate threat mitigation and malicious traffic redirection.HONEYPOT\_REDIRECT action converts the suspicious traffic into an actionable threat intelligence with verified attack success rates above a 99\% in the later rounds.
    \item A four-level graduated degradation strategy is designed to preserve critical IoMT services during cyberattacks with minimal performance impact.
    \item An explainable decision framework is incorporated using SHAP, LIME, Grad-CAM, and counterfactual analysis to improve transparency and accountability.
\end{itemize}

The remainder of this paper is organised as follows. Section~\ref{sec:related} surveys related work. Section~\ref{sec:problem} presents the problem formulation. Section~\ref{sec:framework} details the Federated TGCN-A2C architecture and algorithmic procedure. Section~\ref{sec:experiments} presents experimental results and comparative evaluation. Section~\ref{sec:resilience} analyses cyber resilience properties and privacy guarantees. Section~\ref{sec:xai} presents explainability analysis. Section~\ref{sec:conclusion} concludes and outlines future directions.

\section{Related Work}
\label{sec:related}

This section reviews the six technology domains whose intersection defines the design space of the Federated TGCN-A2C framework, identifying the specific limitations of existing approaches that motivate each design choice in our framework.

\subsection{Digital Twins for IoT and Healthcare Security}

The concept of a Digital Twin is a continuously synchronised virtual model of a physical entity. It was originally proposed for industrial manufacturing. But it has since been adapted broadly to the networked systems security. Alcaraz and L\'opez \cite{alcaraz2022} conducted a foundational survey of the security threats specific to the DT environments, cataloguing the synchronisation latency attacks, the twin state poisoning, and the data-integrity violations as the primary vectors. Their analysis established that a synchronisation delay exceeding approximately the 0.05~ms on a high-fidelity industrial DT is a reliable indicator of man-in-the-middle interception. This finding motivates the use of the reconstruction error as an anomaly signal in our design.

Specifically in the IoT domain specifically, Khan et al.\ \cite{khan2022} proposed a taxonomy of the DT applications for wireless systems. A partitioning of the uses was performed by the authors, yielding three categories: a design validation category, a predictive maintenance category, and a security simulation category. A key observation from their work is that a primary obstacle for a DT deployment on a resource-constrained IoT hardware platform is a memory footprint associated with the twin model itself. They addressed this by offloading a twin computation to the edge gateways. Our Digital Twin design follows this prescription: twin inference executes on the edge gateway rather than the medical device, keeping the device-side overhead to the cost of transmitting compressed state updates.

For the healthcare setting specifically, a growing body of work has explored how DTs can secure clinical environments without exposing raw patient data. Qureshi et al.\ \cite{qureshi2023dt} proposed a federated DT architecture for smart hospitals that separates the learning of device behaviour models from the collection of patient identifiers, and demonstrated that reconstruction-based anomaly detection achieves 93\% detection of ransomware pre-cursors with a false positive rate below 2\%. Chen et al.\ \cite{chen2024} extended this line of work by incorporating blockchain-based attestation into DT synchronisation for medical device authentication, achieving 99.2\% authentication accuracy at 1.5~ms latency. These studies collectively motivate the use of reconstruction-error-based Digital Twins as a first-line anomaly signal, but none of them couple the DT output directly into the classifier inference pathway as a learned gate, a design choice we introduce in the Federated TGCN-A2C framework.

\subsection{Federated Learning for Privacy-Preserving IoMT Security}

Federated learning, formalised by McMahan et al.\ \cite{mcmahan2017communication} as FedAvg, has become the dominant paradigm for privacy-preserving collaborative learning. Its core guarantee, that raw data never leaves the originating device, maps directly onto the data-residency requirements of healthcare regulation and has motivated a substantial body of work applying FL to medical IoT security.

Mothukuri et al.\ \cite{mothukuri2021survey} surveyed FL applications in healthcare, identifying three fundamental challenges: statistical heterogeneity of data across participating institutions (the non-IID problem), communication efficiency in bandwidth-constrained clinical networks, and vulnerability to model-poisoning attacks through malicious gradient updates. Statistical heterogeneity is particularly acute in healthcare: in our experimental setup on CICDDoS~2019, Client~0 carries 47,020 training windows with TFTP as its dominant class (38.7\%), Client~1 carries 46,991 windows with DrDoS\_NTP as dominant (38.6\%), and Client~2 carries 44,020 windows with a more balanced distribution, reflecting realistic hospital network heterogeneity.

Li et al.\ \cite{li2023} developed the enhanced federated frameworks for the heterogeneous IoT that address the statistical heterogeneity through an adaptive per-client weighting, achieving a strong accuracy in health monitoring tasks. Zhang et al.\ \cite{zhang2022fl} applied a transfer learning within a federated framework to the IIoT intrusion detection, achieving the 95.97\% accuracy, though their use of a shared cloud server for the aggregation introduces the 85--120~ms round-trip latency unsuitable for time-critical medical applications. Singh et al.\ \cite{singh2023} proposed a federated IDS for the healthcare IoT, achieving the 98.2\% accuracy on NSL-KDD, though that dataset lacks the contemporary attack vectors relevant to the IoMT threat landscape.

A key algorithmic advance relevant to our work is SCAFFOLD \cite{karimireddy2021scaffold}, which addresses the client drift in the non-IID settings through the control variates that correct the direction of the local gradient updates. SCAFFOLD reduces the communication rounds by up to a 50\% compared to the FedAvg on the heterogeneous benchmarks. Our EMA weighting approach addresses the client drift from a different angle: rather than modifying the gradient update rule, we smooth the aggregation coefficients over time, preventing the instantaneous validation loss spikes from propagating to the global model. This is simpler to implement and does not require the additional communication overhead.

Recent work has also examined the communication cost of the federated learning in the hospital networks specifically. Nguyen et al.\ \cite{nguyen2022federated} studied the federated IDS convergence across the simulated hospital topologies and found that an adaptive client selection can reduce the total bytes transmitted by a 40\% with a minimal accuracy loss. Our cosine annealing of the local learning rate per federated round achieves a similar effect: in the early rounds, when local models diverge rapidly, larger local steps accelerate the convergence; in the later rounds, when the global model is near optimal, the smaller steps prevent destructive local overfitting before aggregation. On CICDDoS~2019, the local learning rate decays from $10^{-3}$ at round~1 to $3.94 \times 10^{-6}$ at round~25.

\subsection{Graph Neural Networks and Temporal Sequence Modelling for Intrusion Detection}

The field of network intrusion detection has undergone a transition from hand-engineered feature extractors toward end-to-end deep learning models that operate directly on raw or lightly processed traffic sequences. Graph Convolutional Networks have established strong baselines by exploiting both feature-level relational structure and the sequential ordering of network flows \cite{nguyen2023survey}. More recently, temporal graph constructions have been combined with GCN layers to allow the model to aggregate information from temporally proximate flow observations within a sliding window.

Zhao et al.\ \cite{zhao2023attention} demonstrated that temporal augmentation of deep architectures achieves significantly better detection of low-rate DDoS attacks, because the model can focus on the sub-second bursts that characterise amplification attack traffic while down-weighting the surrounding baseline. Ullah et al.\ \cite{ullah2024iot} applied a deep IoT traffic classifier to the TON-IoT benchmark, achieving 99.1\% accuracy on nine-class classification, though their work did not incorporate Digital Twin anomaly gating or federated training. Li et al.\ \cite{li2024fedrl} explored the integration of deep classifiers into federated settings for IoT security and found that EMA-weighted aggregation converges more stably than uniform FedAvg when clients have heterogeneous class distributions, a finding that directly motivates our aggregation design.

It is important to note the architectural appropriateness of the Temporal GCN for flow-level datasets. The CICDDoS~2019 and TON-IoT datasets provide pre-computed flow-level statistical features generated by CICFlowMeter; raw packet IP and port identifiers are dropped during feature extraction. The temporal graph is constructed from a $k$-nearest-neighbour edge index over window nodes.
The graph, therefore, captures temporal flow correlations rather than device communication topology. The GNN-based approaches require an explicit device communication graph \cite{nguyen2023survey} and are therefore not applicable to these feature representations without re-processing the original PCAPs. The Temporal GCN operates on ordered windows of flow statistics, and, for these datasets, it is the architecturally appropriate choice.

\subsection{Honeypot Systems and Deception-Based Defense}

A honeypot holds a distinctive niche within the cybersecurity ecosystem: unlike detection systems that passively observe traffic, the honeypots actively interact with attackers, wasting adversary resources while simultaneously harvesting high-fidelity intelligence.
That intelligence covers attack tools, techniques, and procedures. In a systematic review of 47 honeypot deployments, Nawrocki et al.\ \cite{nawrocki2023} quantified that high-interaction honeypots capture, on average $3.2\times$ more sophisticated attacks than low-interaction alternatives.

In the IoT context, honeypots face the additional challenge of emulating the specific protocol stacks and firmware behaviours of medical devices. Gen\c{c} et al.\ \cite{genc2022} developed IoT-POT, a high-interaction honeypot capable of emulating 12~different device types and collected over 4,700 attacks in a 30-day deployment. Huang et al.\ \cite{huang2024honeypot} proposed an adaptive medical device honeypot that adjusts its emulated vulnerability profile based on the attacker's probing behaviour, increasing dwell time from an average of 4.2 minutes for static honeypots to over 18~minutes for adaptive implementations. Kumar et al.\ \cite{kumar2023} integrated honeypot management directly with deep Q-learning, allowing the RL agent to decide in real time whether to redirect suspicious traffic to a honeypot or isolate it outright, achieving a 47\% improvement in attacker engagement time.

Our Enhanced Honeypot System builds on this RL-integration paradigm with adaptive threshold adjustment. The A2C agent's HONEYPOT\_REDIRECT action is selected based on the seven-dimensional state vector that captures both the GCN's classification confidence and the Digital Twin's anomaly score. An adaptive capture threshold $\tau$ begins at 0.70, decreasing by a factor of 0.95 every 10 successful captures (floor 0.40) and increasing by a factor of 1.05 every three failed captures (ceiling 0.90), producing redirect decisions that are substantially more precise than random or threshold-based policies. On the CICDDoS~2019, the thresholds stabilise at a 0.400 by round~4 for all three clients, reflecting a rapid calibration as the GCN's high confidence scores consistently exceed the threshold for the true attacks.

\subsection{Reinforcement Learning for Adaptive Network Defense}

Reinforcement learning has emerged as a natural fit for the network defense. Because the problem structure matches the RL formalism. An agent observes a partially observable system state and selects from a discrete action space. The agent receives the reward signals whose optimisation aligns with the security objectives. Nguyen and Reddi \cite{nguyen2021} surveyed 124 deep RL applications in cybersecurity, identifying the intrusion response as the domain where the RL most clearly outperforms the rule-based baselines, with reported improvements of 15--30\% in the dynamic threat environments.

Among the policy gradient methods applied to network defense, Advantage Actor-Critic provides a stable on-policy baseline that balances the sample efficiency with the convergence stability for the discrete action spaces \cite{schulman2017}. Zhou et al.\ \cite{zhou2023} applied the PPO to DDoS mitigation in the IoT networks and achieved the 96.3\% mitigation effectiveness while maintaining the 94.1\% legitimate traffic throughput. Li et al.\ \cite{li2024fedrl} extended the RL to the federated settings for the industrial IoT, demonstrating that aggregating the actor networks with the uniform weights produces the more stable policies because the reward signal is not directly comparable across the non-IID clients. This finding motivates our design choice to aggregate the  Temporal GCN models with EMA-weighted coefficients while aggregating the A2C models with the uniform weights.

\subsection{Integrated Frameworks for IoMT Security}

Several recent works have combined the multiple technologies toward the comprehensive IoMT protection. Salim et al.\ \cite{salim2022} proposed DTFL-CD, a blockchain-enabled Digital Twin framework that integrates the federated learning with an anomaly threshold scoring mechanism for the early botnet detection in the IIoT environments. DTFL-CD achieved F1 scores of 0.98 on the CICDDoS~2019 but required the 48~federated rounds and exhibited the CPU utilisation between 44\% and 71\%. Its primary limitation is that it does not include an adaptive response component: once an anomaly is detected, the system raises an alert but delegates the response decision to the human operators, limiting its utility in the automated clinical environments.

The closest prior integrated work to our framework is Fed-Inforce-Fusion \cite{khan2022a}, which combined federated learning with Q-learning-based reinforcement for security and privacy protection of IoMT networks. On the TON-IoT dataset, Fed-Inforce-Fusion achieved 99.40\% accuracy and a 98.99\% detection rate across seven attack classes. However, several gaps remain: the Q-learning approach uses a discrete, tabular value function that does not scale gracefully to continuous state spaces; the system does not incorporate Digital Twin monitoring; and there is no honeypot integration. The Federated TGCN-A2C framework addresses each of these gaps directly, achieving 99.48\% on CICDDoS~2019 and 99.61\% on TON-IoT while covering the full ten-class distribution, including the heavily imbalanced MITM class.

\subsection{Explainability in Deep Learning-Based Intrusion Detection}

The deployment of deep learning models in safety-critical healthcare environments necessitates mechanisms that provide human-interpretable evidence for model decisions. SHAP (SHapley Additive exPlanations) \cite{lundberg2017shap} provides theoretically grounded feature attributions based on cooperative game theory, and has been applied to network traffic classification to identify the flow statistics most predictive of attack behaviour. LIME (Local Interpretable Model-Agnostic Explanations) \cite{ribeiro2016lime} approximates the local decision boundary of any classifier with a sparse linear model, enabling per-sample explanations without architectural modification. Grad-CAM \cite{selvaraju2020gradcam}, originally proposed for convolutional networks, can be adapted to GCN-based models by computing gradient-weighted activations over the node feature representations, producing spatial and temporal saliency maps that identify which nodes and features drove a classification. Counterfactual explanation methods \cite{wachter2017counterfactual} identify the minimal feature perturbation required to change a model's prediction, providing contrastive evidence that is particularly useful for auditing false positives and false negatives in clinical settings. The Federated TGCN-A2C framework incorporates all four methods as a post-deployment explainability layer, operating on the trained Temporal GCN without modifying its architecture or inference pipeline.

\section{Problem Formulation}
\label{sec:problem}

We consider an IoT and IoMT deployment consisting of $N$ heterogeneous devices distributed across $K$ geographically separated healthcare facilities, each modelled as an independent federated client. The adversary possesses network access to at least one traffic segment and is capable of launching volumetric flooding, amplification attacks (DNS, NTP, SNMP, LDAP, MSSQL, NetBIOS, UDP), TCP SYN flooding, TFTP exploitation, backdoor establishment, credential brute-forcing, injection attacks, man-in-the-middle interception, ransomware deployment, reconnaissance scanning, and cross-site scripting. The adversary is not assumed to have access to the training process, model parameters, or Digital Twin replicas, but may conduct slow, low-rate attacks designed to evade threshold-based detectors and may adapt tactics over time. The defender's objective is twofold: classify each time window of network flow statistics into the correct class with high accuracy, and select an appropriate response action in real time without disrupting clinical operations.

Network traffic is modelled as a time-ordered sequence of $F$-dimensional flow-level feature vectors. Given observations $\{x_1, x_2, \ldots, x_N\}$ with associated class labels $\ell_i \in \{0, 1, \ldots, C-1\}$, sliding windows of width $W$ and step $S$ are constructed as $W_p = \{x_{pS},\, x_{pS+1},\, \ldots,\, x_{pS+W-1}\}$, with each window assigned the majority class of its constituent samples. For each device $i$, a lightweight Digital Twin $\mathcal{T}_i$ trained on normal-class traffic produces a per-device anomaly score $a_i(t) = \frac{1}{F}\sum_{j=1}^{F}|x_{i,j}^{(t)} - \hat{x}_{i,j}^{(t)}|$ at inference time, providing a complementary signal for detecting novel or stealthy attacks. The Temporal GCN $f_\theta$ maps each window graph to a probability distribution over $C$ classes modulated by $a_i$, from which the A2C response agent observes a seven-dimensional state $s = [p_{\max},\,\min(H,5.0),\,\min(a_i,5.0),\,\mathbf{1}[\ell = c_{\mathrm{normal}}],\,n_k/50000,\,r_{\mathrm{normal}},\,\sigma_{\mathrm{probs}}]$ and selects from three actions: ALLOW, ISOLATE, and HONEYPOT\_REDIRECT. Federated aggregation employs EMA-smoothed per-client validation losses $\ell_k^{\mathrm{EMA}}(r) = \alpha\,\ell_k(r) + (1-\alpha)\,\ell_k^{\mathrm{EMA}}(r-1)$ as inverse weights $w_k = 1/(\ell_k^{\mathrm{EMA}}(r) + \varepsilon)$ to form the global model $\theta_{\mathrm{global}}^r = \sum_{k=1}^{K} (w_k/\sum_j w_j)\,\theta_k^r$, assigning higher weight to clients with lower smoothed validation loss while the $\varepsilon$ term prevents numerical instability.

\section{Proposed Framework Architecture}
\label{sec:framework}

The Federated TGCN-A2C framework is organised into four coordinated layers, Device, Edge, Cloud, and Application, that together provide end-to-end security from individual device monitoring to global threat intelligence.

\begin{figure}[H]
\centering
\includegraphics[width=1.1\textwidth]{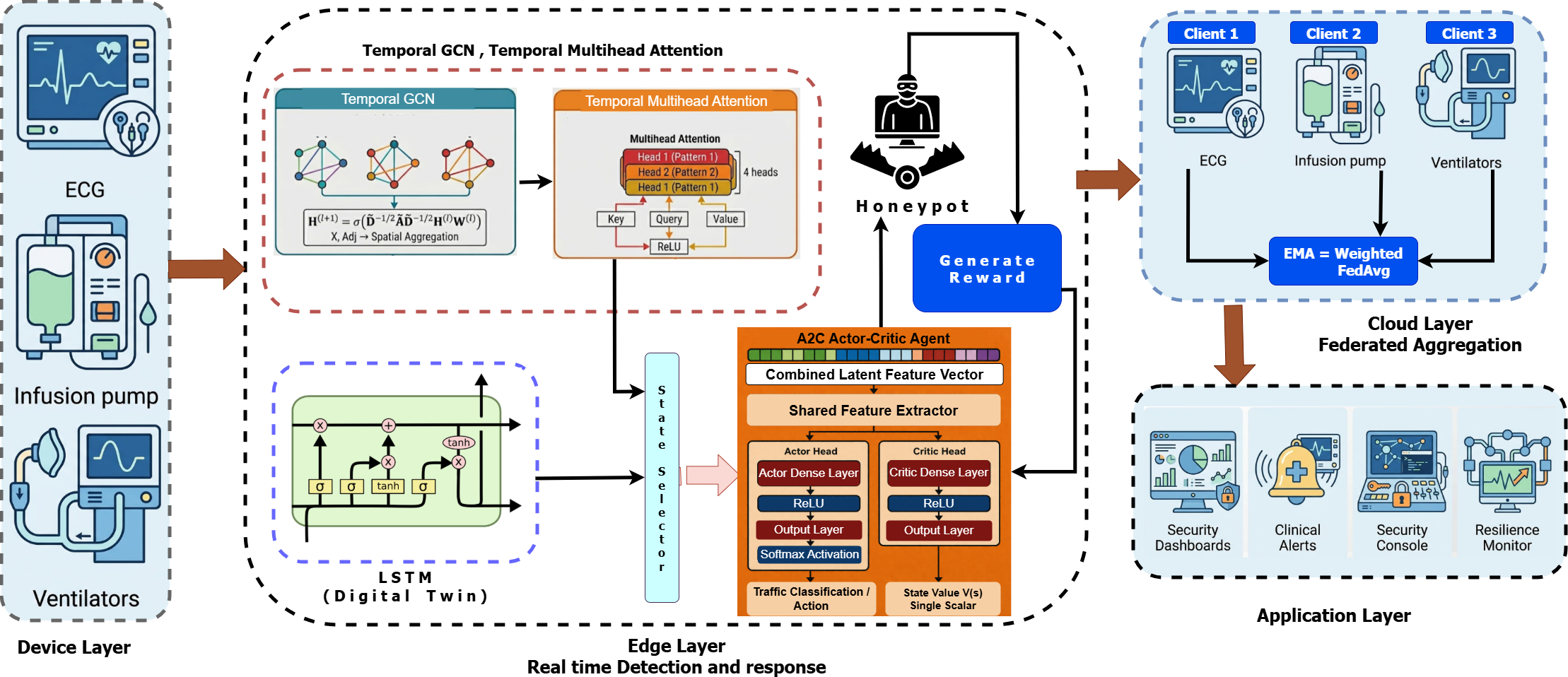}
\caption{Four-layer Federated TGCN-A2C architecture showing Device, Edge, Cloud, and Application layers. The resilience manager overlay in the Edge Layer dynamically reallocates compute between operational processing and security functions across four threat levels (Level~0 to Level~3), ensuring clinical operational continuity at every stage of an active incident.}
\label{fig:architecture}
\end{figure}

Figure~\ref{fig:architecture} illustrates the overall architecture of the Federated TGCN-A2C framework. The Device Layer contains physical IoT and IoMT equipment that transmits encrypted flow statistics to the edge. The Edge Layer hosts the Digital Twins, Temporal GCN classifier, A2C response agent, and Enhanced Honeypot System as the primary computational site for real-time threat detection and response. A resilience manager in the Edge Layer continuously monitors the system threat level and dynamically reallocates edge compute between operational data processing and security functions, ranging from 70/30 at Level~0 up to 10/90 at Level~3, ensuring life-critical device services remain uninterrupted at every threat level. The Cloud Layer performs an EMA-weighted federated aggregation. It distributes the updated global models back to the clients. The Application Layer provides role-appropriate monitoring dashboards for the clinical staff and the security engineers, exposing per-device anomaly scores, honeypot capture rates, A2C action distributions, and federated training progress.
\FloatBarrier

\subsection{Device Layer}

The Device Layer consists of the physical IoT and IoMT endpoints, pacemakers, infusion pumps, monitoring wearables, industrial sensors, and the gateway devices. Each device transmits encrypted flow-level statistics to the edge gateway over TLS~1.3 with certificate-based mutual authentication and perfect forward secrecy. Patient raw data (e.g., ECG waveforms, glucose readings) never leaves the device at any point. Only pre-computed network flow features extracted by an on-device lightweight agent are transmitted. This design ensures that even a fully compromised edge gateway cannot reconstruct the measurements of an individual patient.

\subsection{Edge Layer: Digital Twin, Temporal GCN, A2C, and Enhanced Honeypot}

The Edge Layer is the primary computational site for real-time threat detection and response. It hosts four components per deployed client.

\subsubsection{Lightweight Digital Twin}

Each device $i$ has a corresponding Digital Twin implemented as a single-layer LSTM with 64~hidden units. The twin is trained exclusively on windows whose majority label is the normal class, learning the device's expected temporal traffic pattern under benign conditions. Training minimises the mean squared reconstruction error on the final-step feature vector. On CICDDoS~2019, the training on 23,982 normal-class windows converges with a loss declining from 0.396 at epoch~5 to 0.055 at epoch~70. This confirms that the LSTM captures the statistical regularity of the benign device traffic with high fidelity. A lower final loss indicates well-separated normal behaviour. On TON-IoT, the training perform on 10,667 normal-class windows. The training converges similarly. This produces a discriminative anomaly signal. The signal  meaningfully separates attack windows from normal windows at inference time. At inference time, the anomaly score is computed for every incoming window and used both as an A2C state component and as input to the Temporal GCN anomaly gate.

\subsubsection{Temporal Graph Convolutional Network}

The Temporal GCN is the primary classification component. It processes windows as graphs where each of the $W$ time steps is a node, and edges are defined by a precomputed $k$-nearest-neighbour temporal edge index with $k = 5$. A linear projection first maps each node from feature space to a $H=128$ dimensional hidden space with ReLU activation. Two GCNConv layers from PyTorch Geometric, then propagate node features over the temporal graph with dropout applied after each layer, implementing the normalised Laplacian aggregation $H' = \tilde{D}^{-1/2}\tilde{A}\tilde{D}^{-1/2}HW$. A global mean pool reduces node-level representations to a graph-level embedding.

The Digital Twin anomaly score $a_i$ enters the GCN through a learned per-class sigmoid gate: $\hat{y} = f_\text{cls}(h_\text{graph}) \cdot (1.0 + g(a_i))$, where $f_\text{cls}$ is a three-layer MLP and $g$ is a two-layer MLP with sigmoid output producing a per-class scaling factor. When the Digital Twin anomaly score is near zero (normal-like traffic), the gate has no effect on the logits. When the anomaly score is large (anomalous traffic), the gate amplifies attack-class logits, increasing classification confidence on the attack side and reducing the likelihood of false negatives on novel attack variants whose flow statistics are ambiguous without the anomaly context.

The Temporal GCN is trained with cross-entropy loss with label smoothing, AdamW optimiser, and cosine-annealed learning rate per federated round. Class-balanced training is ensured through a WeightedRandomSampler that assigns sample weights inversely proportional to class frequency in each client's local partition, providing equal expected exposure for all classes per epoch, including heavily underrepresented classes such as \textit{mitm}.

\subsubsection{A2C Response Agent}

The A2C agent is a two-head neural network with a shared feature extractor: two linear layers with 128~hidden units and ReLU activations, followed by an actor head that outputs action logits and a critic head that produces scalar value estimates. Weights are initialised with Xavier uniform initialisation.

During experience collection, the Temporal GCN is set to evaluation mode to stabilise the state representations. Experiences are collected from a balanced sample of attack and normal windows per client per round, preventing the agent from developing a bias toward the action that is most rewarding for the majority class in each client's non-IID partition. For the A2C update, we rely on the standard actor-critic objective. A setting of 0.01 is used for an entropy regularisation coefficient. A discount factor $\gamma = 0.99$ is employed, and a gradient clipping is applied at a norm of 1.0. A reward function operates according to the following rules. A correct HONEYPOT\_REDIRECT action for an attack receives a base reward of 4.0 together with a confidence bonus and an anomaly bonus. A correct ALLOW action for a normal traffic receives a reward of 2.5 plus a confidence bonus. A correct ISOLATE action for an attack receives a reward of 2.0 plus a confidence bonus. A penalty is applied to any incorrect action instead of a reward. A decay of an exploration rate occurs from an initial value of 0.30 down to a minimum value of 0.05 per round. This gradual rate of decay ensures a sufficient level of early exploration across an action space.

\subsubsection{Enhanced Honeypot System}

Enhanced Honeypot System: An interception of traffic segments is performed by this system. For each intercepted segment, a HONEYPOT\_REDIRECT action is selected by the A2C agent. A maintenance of cumulative statistics is also performed by the system. These statistics include a total number of attempts, a number of successful captures, a number of failed captures, an attack type distribution, an average confidence value, and a per-round performance history.There is an adaptive capture threshold $\tau$ that begins at 0.70. It decreases by a factor of 0.95 every 10 successful captures (floor 0.40), and it increases by a factor of 1.05 every three failed captures (ceiling 0.90). On the CICDDoS~2019 dataset, the thresholds stabilise at 0.400 by round~4 for all three clients. These results reflect rapid calibration as the GCN's high confidence scores consistently exceed the threshold for true attacks from early rounds onward. The Enhanced Honeypot System logs full feature vectors. It logs GCN probability distributions, Digital Twin anomaly scores, and timestamps for each captured event. In this way, it populates a growing threat intelligence database.

\subsection{Cloud Layer: Federated Aggregation and Global Intelligence}

At the Cloud Layer, client model updates are aggregated. It uses EMA-weighted FedAvg for the Temporal GCN, and it uses a uniform FedAvg for the A2C agents. For the GCN, aggregation weights are computed as the inverse of the EMA-smoothed validation loss: $w_k^r = (1/(\ell_k^{\mathrm{EMA}}(r) + \varepsilon)) / \sum_{j=1}^K (1/(\ell_j^{\mathrm{EMA}}(r) + \varepsilon))$, with $\varepsilon = 10^{-6}$. A2C models are aggregated with uniform weights. The reason for this difference is that return estimates under different non-IID reward distributions are not directly comparable across clients. The server then evaluates the global Temporal GCN on the held-out validation and test partitions. It saves a checkpoint whenever the validation accuracy improves. Early stopping is activated when the 5-round rolling mean validation F1 change falls below $3 \times 10^{-4}$ and also activates when the mean exceeds 0.97. This occurs after round~19. Additionally, the Cloud Layer also aggregates honeypot capture logs from all clients, thereby building a global threat intelligence database.

\subsection{Application Layer}

At the Application Layer, the system's security posture is exposed through a web-based dashboard that provides a real-time visualisation of per-device anomaly scores, honeypot capture rates and precision, A2C action distributions, and federated training progress. Role-appropriate interfaces are provided for clinical staff and security engineers. Traffic at the Application Layer is isolated from clinical device communication networks and authenticated through role-based access control.

\FloatBarrier

\subsection{Complete Framework Algorithm}

A complete procedure for the Federated TGCN-A2C framework is presented in Algorithm~\ref{alg:main}. The procedure is organized as a set of three sequential phases: an offline initialisation phase, a federated training phase, and a live deployment phase. A guarantee of operational continuity applies to all three phases.

\begin{algorithm}
\caption{Federated TGCN-A2C: Main Federated Training and Inference Workflow}
\label{alg:main}
\begin{algorithmic}[1]
\scriptsize
\Require $K$ clients with local data, $R$ federated rounds, EMA factor $\alpha=0.4$, early stopping threshold 0.97
\Ensure Trained global classifier, response policy, and threat database

\Statex \textbf{Phase 1: Initialization}
\State Build a temporal $k$-NN edge index with $k=5$ for the sliding window size $W$ and share it across all clients to avoid redundant graph construction during training
\State Train a Digital Twin model for each client using only normal traffic samples until the reconstruction error stops decreasing significantly
\State Initialize the global Temporal GCN and A2C policy networks with Xavier uniform weights
\State Deploy Enhanced Honeypot Systems for each client with an initial adaptive capture threshold of $\tau = 0.70$
\State Set up Exponential Moving Average loss trackers for every client, initially set to zero

\Statex \textbf{Phase 2: Federated Training (Repeated for $R$ rounds)}
\For{each federated round $r = 1$ to $R$}
    \State Compute the cosine-annealed learning rate for this round as $\mathrm{lr}_r = \mathrm{lr}_0 \times 0.5 \times (1 + \cos(\pi r / R))$
    \For{each client $k = 1$ to $K$ in parallel}
        \State Download the latest global Temporal GCN and A2C models from the central server
        \State Train the Temporal GCN locally on the client's data for 70 epochs using class-balanced sampling and the anomaly gate driven by Digital Twin reconstruction error
        \State Compute the validation loss on the client's held-out set and update its EMA-smoothed loss value using the factor $\alpha = 0.4$
        \State Switch the Temporal GCN to evaluation mode to collect stable A2C training experiences
        \State Build a balanced experience buffer containing equal numbers of attack and normal traffic samples
        \State For each experience, compute the Digital Twin anomaly score, construct the seven-dimensional state vector, sample an A2C action with a decayed exploration rate, and compute the structured reward
        \State If the selected action is HONEYPOT\_REDIRECT and the traffic is a confirmed attack, engage the Enhanced Honeypot and adjust the adaptive threshold based on capture success
        \State Perform the A2C gradient update with entropy regularisation and clip the gradient norm to 1.0
        \State Upload the updated local Temporal GCN weights, A2C weights, and EMA validation loss back to the server
    \EndFor
    \State Calculate aggregation weights for each client as the inverse of their EMA validation loss, adding a small constant $\varepsilon$ to avoid division by zero
    \State Normalize these weights so they sum to one across all clients
    \State Aggregate all client Temporal GCN models into a new global model using the normalized weights, giving higher influence to clients with lower validation loss
    \State Aggregate all client A2C models into a new global A2C model using uniform averaging with equal weight for every client
    \State Evaluate the new global Temporal GCN on the central validation and test sets
    \If{validation accuracy has improved compared to the previous best value}
        \State Save the current global Temporal GCN and A2C models as the new best checkpoint
    \EndIf
    \If{round $\geq 19$ and the 5-round rolling F1 change is below $3\times10^{-4}$ and the mean F1 exceeds 0.97}
        \State \textbf{break}
    \EndIf
\EndFor

\Statex \textbf{Phase 3: Real-Time Inference}
\While{the system is operational}
    \State For each active device, compute an anomaly score using its trained Digital Twin by measuring reconstruction error on the latest traffic window
    \State Pass the recent traffic window graph through the Temporal GCN to obtain probability scores for each possible attack class, modulated by the anomaly gate
    \State Build a seven-dimensional state vector containing: highest class probability, entropy of the probability distribution, Digital Twin anomaly score, normal traffic indicator, normalized client sample count, normal traffic ratio, and standard deviation of class probabilities
    \State Feed this state vector to the trained A2C policy network to select one of three possible actions: forward traffic normally, quarantine the suspicious device, or redirect traffic to the honeypot for intelligence gathering
    \State Execute the selected action; if the action is HONEYPOT\_REDIRECT, log the full traffic features, GCN probabilities, anomaly score, and timestamp to the threat intelligence database
    \State Reassess threat level and reallocate edge resources per the four-level degradation policy.
\EndWhile
\end{algorithmic}
\end{algorithm}
\newpage

\subsection*{}

The Federated TGCN-A2C algorithm is organized into three sequential phases that together provide a complete cyber-resilient security solution for IoT and IoMT networks.

\textbf{Phase 1 (Lines 3--7) -- Initialization:} This phase executes only once before any federated training begins. Line 3 precomputes the temporal $k$-NN edge index shared across all windows, to avoide redundant graph construction during training. Then line 4 trains a lightweight LSTM-based Digital Twin for each client. A key detail: training happens exclusively on normal-class windows. This establishes a behavioural baseline for each device without requiring any labelled attack data. The twin is trained until reconstruction loss plateaus, confirming that the LSTM captures the statistical regularity of benign traffic with high fidelity. Line 5 initialises the global Temporal GCN and A2C policy networks with Xavier uniform weights, which will be improved through federated training. Line 6 deploys Enhanced Honeypot Systems with an initial adaptive capture threshold $\tau = 0.70$. Line 7 sets up EMA loss trackers for every client, initially set to zero, which will smooth validation losses across rounds to prevent sudden fluctuations from destabilising the aggregation weights.

\textbf{Phase 2 (Lines 9--27) -- Federated Training:} This phase executes for up to $R$ federated rounds. At the beginning of each round, the cosine-annealed learning rate is computed (Line 10) and applied to all client GCN optimisers. Every client in parallel downloads the latest global models (Line 12). Line 13 performs local Temporal GCN training for 70 epochs using class-balanced WeightedRandomSampler and the anomaly gate that amplifies attack-class logits when the Digital Twin detects anomalous behaviour. After local training, Line 14 computes the validation loss and updates the EMA-smoothed loss with factor $\alpha = 0.4$, giving 40\% weight to the current round's loss and 60\% to the historical average. Line 15 switches the GCN to evaluation mode and collects 200 balanced A2C experiences. A seven-dimensional state is constructed at Line 16. This state draws from four sources: a set of GCN output probabilities, an entropy value, a Digital Twin anomaly score, and a set of traffic composition statistics. A sampling of actions occurs at the same line using a decayed exploration mechanism. Following this, a computation of structured rewards is performed. A honeypot engagement occurs at Line 17 for a purpose of confirmed attack re-directions, and an adjustment of the adaptive threshold takes place based on a capture success status. A gradient update for the A2C is performed at Line 18. A set of local models together with an EMA loss value is uploaded to the server at Line 19.

Lines 20--27 perform server-side aggregation and evaluation. Line 20 computes EMA-inverse aggregation weights. Clients with lower validation loss receive a stronger influence in the global GCN through these weights. Line 21 normalises the weights. Line 22 aggregates GCN models with EMA weights. Line 23 aggregates A2C models with uniform weights. Since reward distributions are not directly comparable across non-IID client partitions. Line 24 evaluates the global model on the validation and test sets. Line 25 saves a checkpoint when validation accuracy improves. Lines 26--27 implement early stopping. Early stopping activates once the rolling F1 has converged and the rolling f1 has also exceeded the 0.97 threshold for five consecutive rounds.

\textbf{Phase 3 (Lines 29--36) -- Real-Time Inference:} This phase runs continuously once training has completed. Starting with line 30: it computes an anomaly score for each active device. How? By running the device's latest traffic window through the trained Digital Twin and then measuring the reconstruction error. A higher score means the device's behaviour has deviated significantly from the learned normal baseline. Line 31 passes the same traffic window graph through the Temporal GCN, with the anomaly gate modulating the output logits. Line 32 constructs the seven-dimensional state vector that captures the current security situation through seven components: the first is  highest class probability (representing confidence in the most likely classification), the second is entropy of the probability distribution (uncertainty), the third is anomaly score from the Digital Twin, the fourth is a binary indicator of whether is traffic appears normal, the fifth is normalised sample count for this client, the sixth is the ratio of normal traffic seen by this client, and seventh is the standard deviation of class probabilities. Line 33 feeds this state vector to the trained A2C policy network. Network outputs a categorical distribution over three possible actions. The action with the highest probability gets selected. Line 34 executes the selected action: ALLOW forwards traffic normally to its destination, ISOLATE quarantines the suspicious device, and it also activates redundant communication paths to maintain clinical operations, and the HONEYPOT\_REDIRECT silently diverts the traffic to the Enhanced Honeypot System for forensic analysis. For a given honeypot redirect event, a logging of several information pieces to a threat intelligence database is performed by the system. These pieces include a set of full traffic features, a set of GCN probabilities, an anomaly score value, and a timestamp. A movement to line number 35 then triggers an update of the overall system threat level and such an update is based on a sequence of recent actions, and a reallocation of edge computing resources follows according to the graduated degradation policy. Under this policy, a Level 0 receives a 70/30 split, a 70\% allocation for operational processing and a 30\% allocation for security. A Level 1 uses a 50/50 split, while a Level 2 shifts toward a 30\% portion for operational processing and a 70\% portion for security. Level 3 goes to 10\% operational processing and 90\% security. This graduating degradation ensures that life-critical medical functions always receive sufficient computational resources even during the active attacks.

\subsection{Time Complexity Analysis}

The computational complexity of the Federated TGCN-A2C framework scales as $O(R \cdot E \cdot N \cdot W \cdot (kH + H^2))$ for the dominant Temporal GCN training component, where $R$ is the number of federated rounds, $E$ is the number of local epochs per round, $N$ is the number of training windows per client, $W$ is the sliding window length, $k$ is the graph neighbourhood size, and $H = 128$ is the GCN hidden dimension. The $kH$ term arises from GCNConv neighbourhood aggregation over $k$-nearest temporal neighbours and the $H^2$ term from the hidden-to-hidden transformation at each GCN layer. The anomaly gate adds a smaller term $O(N \cdot H \cdot C)$ for per-class sigmoid scaling. Digital Twin training incurs a one-time offline cost of $O(E_{dt} \cdot N_\text{normal} \cdot W \cdot h_{dt} \cdot F)$, where $E_{dt}$ is the number of twin training epochs, $N_\text{normal}$ is the count of normal-class windows, $h_{dt} = 64$ is the LSTM hidden dimension, and $F$ is the input feature dimension. The A2C update contributes $O(B_\text{a2c} \cdot H_\text{a2c}^2)$ per round, where $B_\text{a2c} = 200$ experiences and $H_\text{a2c} = 128$. On the server, federated aggregation requires $O(K \cdot P)$ per round, where $K$ is the number of clients, and $P$ is the GCN parameter count, a cost that is negligible compared to local training. The $K$ clients operate in parallel, reducing wall-clock time by a factor of $K$. For real-time inference on a single window, the Temporal GCN requires $O(W \cdot (kH + H^2))$ for GCNConv propagation and global pooling, The Digital Twin requires $O(W \cdot h_{dt} \cdot F)$, and the A2C decision adds $O(H_\text{a2c}^2)$. Under the deployed hyperparameter settings ($H = 128$, $W = 25$--$40$, $F = 16$--$65$), the total inference latency remains under 50 milliseconds. This comfortably satisfies the real-time requirements of intensive care unit monitoring systems.

\FloatBarrier

\section{Experimental Evaluation}
\label{sec:experiments}

A presentation of the experimental setup, a set of datasets, a description of training dynamics, and a collection of comparative results for the Federated TGCN-A2C framework is provided in this section. A description of two benchmark datasets together with their preprocessing pipelines serves as a starting point. An extension of the coverage then follows toward a federated training progression, a per-class performance analysis, an evaluation of honeypot effectiveness, and a comparison against a set of state-of-the-art baselines.

\subsection{Datasets}

\subsubsection{CICDDoS~2019 Dataset}

CICDDoS~2019 dataset \cite{cicddos2019}: This dataset was created by the Canadian Institute for Cybersecurity. It represents one of the most comprehensive publicly available benchmarks for DDoS detection research. It captures the network traffic from a controlled testbed environment, where a traffic generator produces 18 labelled classes within that environment. These classes include a benign class and 17 distinct DDoS attack categories, covering both reflection/amplification and direct exploitation techniques. Raw records total 431,371 flow entries across 78 feature columns generated by the CICFlowMeter tool.

After preprocessing, we removed columns with missingness exceeding 50\%, dropping identifier fields, coercing all remaining columns to numeric types, and applying column-wise median imputation for residual missing values. After these steps, 65 features were retained. Extreme values were clipped to standard deviation bounds, and per-feature standardisation was fitted exclusively on the training split to prevent data leakage. Sliding windows with a width of 25 and a step of 2 produced 215,674 temporal windows distributed across 16 classes. The two smallest raw classes had insufficient samples for majority-vote window labelling, so they do not appear in the final windowed dataset. A stratified 64\%/16\%/20\% train/validation/test split yields 138,031, 34,508, and 43,135 windows, respectively. The Digital Twin was trained on 23,982 normal-class training windows. For the three non-IID client partitions, we obtained 47,020, 46,991, and 44,020 training windows, respectively. Client~0 carries TFTP as its dominant class (38.7\%). Client~1 carries DrDoS\_NTP as its dominant class (38.6\%), and Client~2 carries a more balanced distribution.

\subsubsection{TON-IoT Dataset}

TON-IoT Dataset \cite{toniot2020}: Developed at UNSW Canberra the Ton-IoT dataset is a next-generation telemetry benchmark. The benchmark specifically evaluates AI-based security solutions in heterogeneous IoT/IIoT environments. Unlike older IDS benchmarks that capture only the network-layer features, TON-IoT integrates three data types. These are OS-level telemetry, IoT sensor readings, and network flow statistics. The version used in this work contains 211,043 records spanning 10 classes. These classes include normal traffic and nine attack categories. The MITM category has only 1,043 records, creating severe class imbalance. This severe class imbalance of the MITM category provides a stringent test of the framework's ability to learn from underrepresented attack types under the federated non-IID partitioning.

After preprocessing and feature selection, the 16 features were retained. Sliding windows with a width of 40 and a step of 3 produce 70,335 temporal windows. The window construction parameters are chosen to match the temporal extent of dominant attack signatures: shorter windows suffice for volumetric flooding attacks, which manifest within the sub-second bursts, while the longer windows are required for behavioural attacks. Behavioral attacks include ransomware, scanning, and credential stuffing. These attacks develop over an extended interaction sequence. A stratified 60\%/20\%/20\% split yields 45,014, 11,254, and 14,067 windows, with the three non-IID clients receiving approximately 15,858, 14,578, and 14,578 training windows. The Digital Twin was trained on 10,667 normal-class training windows.

\subsection{Experimental Setup}

All experiments were conducted on Kaggle's GPU environment using PyTorch with PyTorch Geometric (PyG) for graph neural network operations. The complete hyperparameter configuration is summarised in Table~\ref{tab:setup}. For CICDDoS~2019, the Temporal GCN uses 2 GCNConv layers with hidden dimension 128, an initial learning rate of $10^{-3}$ for Clients~0 and~2 and $10^{-4}$ for Client~1, with cosine annealing per round and AdamW with weight decay $10^{-5}$. Client~1 uses a lower learning rate to reflect its different local distribution and smaller relative influence. Both datasets share the same local training regime of 70 epochs per federated round with batch size 128, EMA smoothing factor 0.4, and A2C hyperparameters including discount factor 0.99, entropy coefficient 0.01, and gradient clip norm 1.0. Label smoothing $\epsilon = 0.02$ is applied during cross-entropy loss computation to prevent overconfidence after federated aggregation, and a WeightedRandomSampler ensures class-balanced mini-batches throughout training. A scheduling of the local learning rate per round is handled by a cosine annealing mechanism. Larger steps are taken during early rounds to achieve a rapid convergence, while smaller steps are used in later rounds for a prevention of destructive overfitting before an aggregation occurs.

\begin{table}[H]
\centering
\caption{Experimental Configuration Parameters for Both Datasets}
\label{tab:setup}
\resizebox{\textwidth}{!}{%
\begin{tabular}{p{0.40\linewidth}p{0.26\linewidth}p{0.26\linewidth}}
\toprule
\textbf{Parameter} & \textbf{CICDDoS~2019} & \textbf{TON-IoT} \\
\midrule
Raw records / total windows & 431,371 / 215,674 & 211,043 / 70,335 \\
Features retained & 65 & 16 \\
Attack classes & 16 & 10 \\
Train / Val / Test windows & 138,031 / 34,508 / 43,135 & 45,014 / 11,254 / 14,067 \\
Client partition sizes & 47,020 / 46,991 / 44,020 & \textasciitilde15,858 / 14,578 / 14,578 \\
Window size $W$ / Step $S$ & 25 / 2 & 40 / 3 \\
Graph $k$-NN neighbours & 5 & 5 \\
Number of federated clients & 3 & 3 \\
Maximum federated rounds & 25 (60 max) & 13 (60 max) \\
GCN hidden dim $H$ & 128 & 128 \\
GCN layers & 2 (GCNConv) & 2 (GCNConv) \\
GCN learning rate (C0/C2) & $1\times10^{-3}$ & $1\times10^{-3}$ \\
GCN learning rate (C1) & $1\times10^{-4}$ & $1\times10^{-4}$ \\
GCN LR schedule & Cosine annealing & Cosine annealing \\
GCN optimiser & AdamW, $\lambda=10^{-5}$ & AdamW, $\lambda=10^{-5}$ \\
Label smoothing $\epsilon$ & 0.02 & 0.02 \\
Local epochs per round & 70 & 70 \\
Local batch size & 128 & 128 \\
GCN gradient clip norm & 1.0 & 1.0 \\
A2C state dimension & 7 & 7 \\
A2C action space & 3 & 3 \\
A2C learning rate (C0/C2) & $3\times10^{-5}$ & $3\times10^{-5}$ \\
A2C learning rate (C1) & $1.5\times10^{-5}$ & $1.5\times10^{-5}$ \\
A2C discount factor $\gamma$ & 0.99 & 0.99 \\
A2C entropy coefficient & 0.01 & 0.01 \\
A2C gradient clip norm & 1.0 & 1.0 \\
EMA smoothing factor $\alpha$ & 0.4 & 0.4 \\
Digital Twin LSTM hidden units & 64 & 64 \\
Digital Twin training epochs & 70 & 70 \\
Digital Twin final loss & $5.5\times10^{-2}$ & converged \\
Honeypot initial threshold $\tau$ & 0.70 & 0.70 \\
Learning rate schedule & Cosine annealing & Cosine annealing \\
\bottomrule
\end{tabular}}
\end{table}

\begin{figure}[H]
\centering
\includegraphics[width=1.1\textwidth]{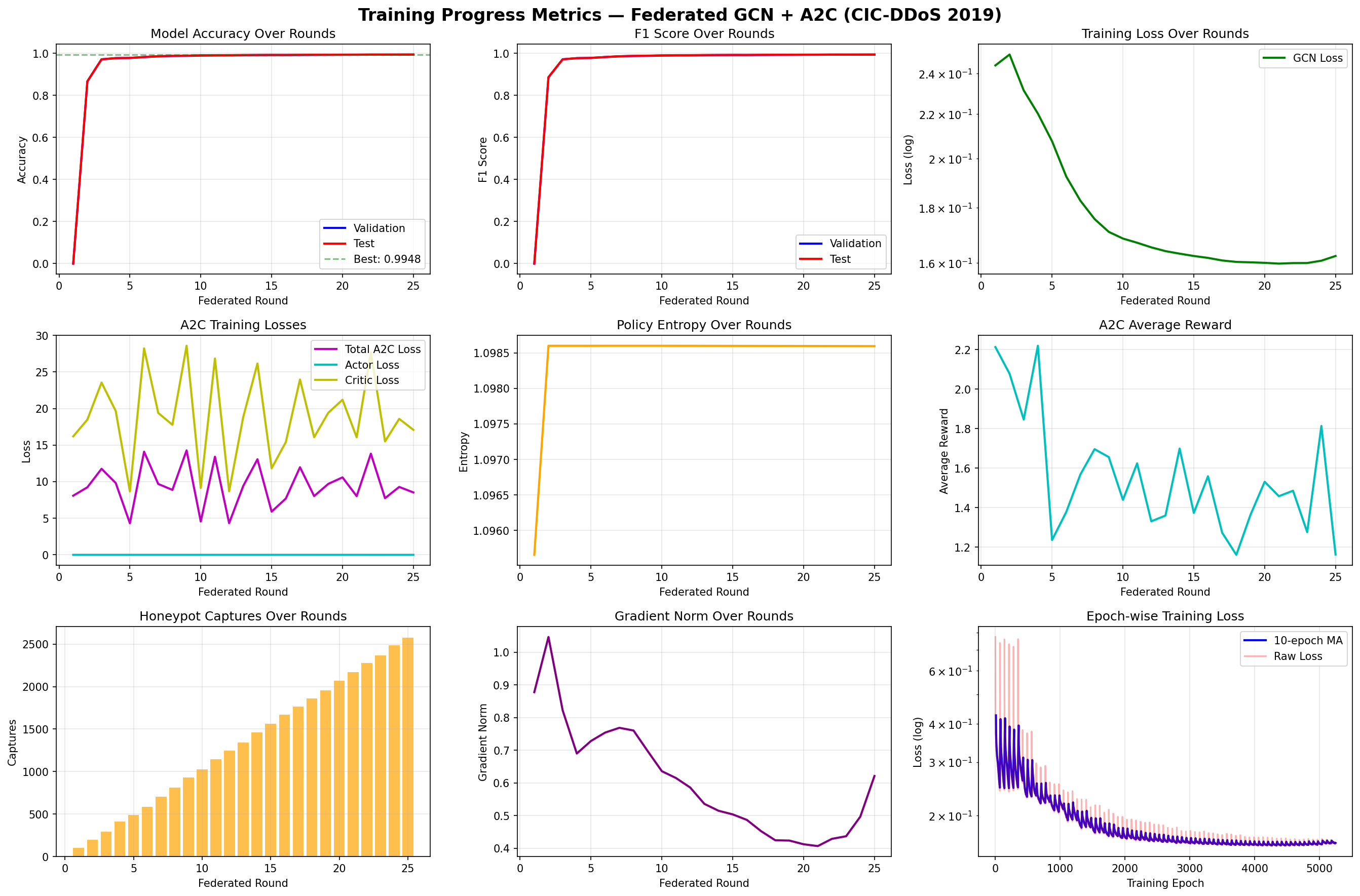}
\caption{Test accuracy, weighted-F1, macro-F1, training loss, honeypot captures, and EMA weight progression across federated rounds for CICDDoS~2019.}
\label{fig:accuracy_cicddos}
\end{figure}

Figure~\ref{fig:accuracy_cicddos} presents the comprehensive training progression on CICDDoS~2019 across six performance dimensions. Looking first at test accuracy, we see rapid convergence across training rounds. Round 1 Starts effectively 0\% before any global aggregation happens, each client's independently trained model can only predict its own dominant class. But after the first global aggregation, the accuracy jumps sharply to 86.77\% at round~2, then improves further to 97.17\% at round 3, reaches 97.77\% at round 4,  and keeps climbing steadily afterward until it reaches 99.48\% at round 25. This monotonic improvement after round~2 reflects the EMA-weighted aggregation strategy, which client weights stable between 0.330--0.337 range for the entire training process. Now consider the macro-F1 score. Its path looks similar to test accuracy: Starting from 0.0000 at round~1 to 0.6237 at round~2, that initial rise reflects trouble with minority classes such as DrDoS\_DNS and DrDoS\_NetBIOS. The score then recovers strongly. By round~3, the score reaches 0.9133 and reached 0.9840 by round~25, while the weighted-F1 score achieved 0.9948 at round~25, which confirms strong performance across all 16 DDoS categories. The GCN training loss declines steadily, from approximately 0.244 per round at round~1 to below 0.162 by round~25. This reflects the convergence of the local GCN models. The honeypot captures increase monotonically from 99 at round~1 to 33,521 by round~25. In later stages, captures accumulate at approximately 1,300--1,500 per round as the GCN's high confidence scores enable consistent redirection decisions. Adaptive thresholds for the honeypot converge from 0.700 to 0.400 by round~4 for all three clients and remain there for the remaining 21 rounds. Finally, EMA aggregation weights stay stable in the 0.330--0.337 range across all 25 rounds, eliminating the weight oscillations that you would normally see with raw inverse-loss weighting.

\begin{figure}[H]
\centering
\includegraphics[width=1.1\textwidth]{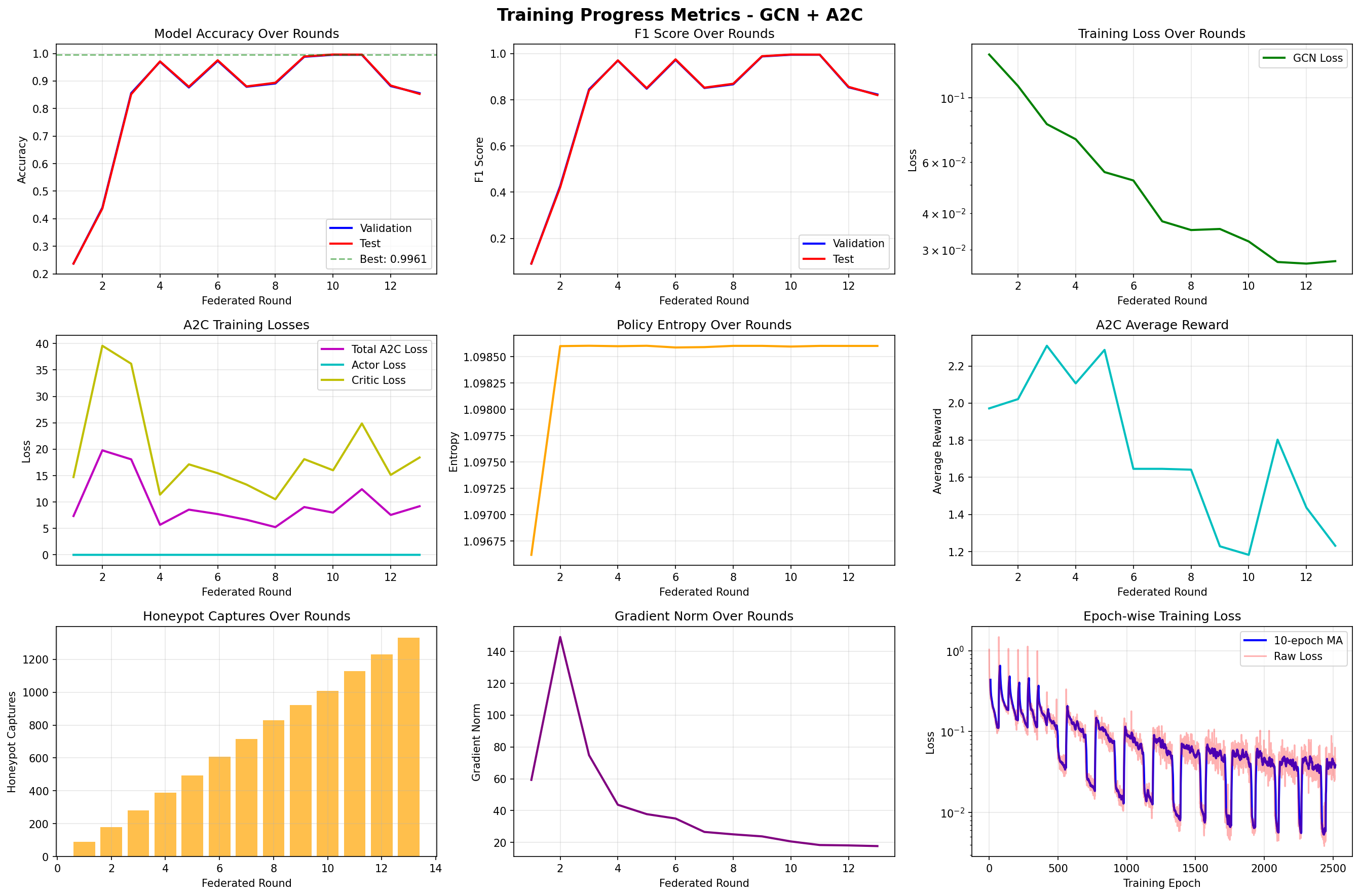}
\caption{Test accuracy, weighted-F1, macro-F1, training loss, honeypot captures, and EMA weight progression across federated rounds for TON-IoT.}
\label{fig:accuracy_toniot}
\end{figure}

Figure~\ref{fig:accuracy_toniot} illustrates the more complex training dynamics on TON-IoT across the same six performance dimensions, revealing the framework's resilience to non-IID client drift. The test accuracy and macro-F1 curves exhibit a distinctive pattern: slow initial convergence occurs, starting from 23.69\% accuracy and a macro-F1 of 0.0383 at round~1. A significant improvement then follows, reaching 85.19\% and 0.7890 by round~3. The curves cross a critical convergence threshold at round~9, where accuracy reaches 98.91\%, and macro-F1 achieves 0.9805. This reflects successful learning of discriminative boundaries for all ten classes include the severely underrepresented MITM category. Performance reaches 99.61\% test accuracy and a macro-F1~0.9912 by round~10. However, late-round oscillations occur at rounds~12--13. These oscillations arise when one client's exceptional local model influences aggregation asymmetrically, and subsequent local updates from other clients pull decision boundaries toward their own partition distributions. The EMA mechanism partially dampens this effect. The best model checkpoint from round~10 is used for the final evaluation. Honeypot captures increase monotonically from 90 at round~1 to 9,206 across the full training run, with the accumulation rate accelerating after round~5. This acceleration occurs as the GCN's improved classification enables more confident redirection decisions. The consistently high weighted-F1 stays above 0.99 from round~5 onward. This remains true despite the late-round oscillations, confirming that the majority of classes maintain strong classification throughout.

\subsection*{Summary of Training Dynamics}

When we examine the six-dimensional performance analysis across both datasets, several key characteristics of the Federated TGCN-A2C framework emerge. First, the EMA-weighted FedAvg aggregation strategy successfully stabilises global model updates under non-IID client distributions on CICDDoS~2019. Evidence for this includes monotonic improvement after round~2 and stable EMA weights in the 0.330--0.337 range. On TON-IoT, stabilisation is partially effective: strong performance is achieved by round~9--10, but late-round oscillations occur due to a smaller dataset and more extreme class imbalance. Second, Temporal GCN with anomaly-gated classification achieves strong convergence within 10--25 federated rounds on both datasets. Test accuracy reaches 99.48\% on CICDDoS~2019 and 99.61\% on TON-IoT. Third, the Digital Twin training is efficient. Twin loss decreases from 0.396 at epoch~5 to 0.055 at epoch~70 on CICDDoS~2019. This provides a well-calibrated anomaly signal. Fourth, the honeypot system accumulates 33,521 and 9,206 verified captures on the two datasets, respectively. Client-level attack success rates consistently above 99\% in later rounds. Finally, the WeightedRandomSampler combined with label smoothing provides effective minority class coverage: all 16 CICDDoS~2019 classes achieve F1~$\geq$~0.9237 and all 10 classes of TON-IoTachieve F1~$\geq$~0.9488, including the severely underrepresented MITM class. This validates the integrated design of the Federated TGCN-A2C framework, where improvements in the GCN classifier propagate to A2C policy effectiveness and honeypot intelligence accumulation.

\FloatBarrier

\subsection{CICDDoS~2019 Training Results}

Table~\ref{tab:cicddos_progress} presents the complete round-by-round training progression on CICDDoS~2019 at selected rounds.

\begin{table}[H]
\centering
\caption{CICDDoS~2019 Federated Training Progress (Selected Rounds)}
\label{tab:cicddos_progress}
\begin{tabular}{cccccc}
\toprule
\textbf{Round} & \textbf{Val Acc} & \textbf{Test Acc}
               & \textbf{Test Macro F1} & \textbf{HP Captures}
               & \textbf{EMA Wt Range} \\
\midrule
 1 & 0.0001 & 0.0000 & 0.0000 &     99 & [0.336, 0.335] \\
 2 & 0.8668 & 0.8677 & 0.6237 &    200 & [0.333, 0.336] \\
 3 & 0.9721 & 0.9717 & 0.9133 &    294 & [0.332, 0.337] \\
 4 & 0.9776 & 0.9777 & 0.9395 &    410 & [0.333, 0.337] \\
 5 & 0.9784 & 0.9786 & 0.9416 &    491 & [0.332, 0.337] \\
 7 & 0.9863 & 0.9865 & 0.9605 &    705 & [0.334, 0.335] \\
10 & 0.9899 & 0.9909 & 0.9717 &  1,027 & [0.334, 0.335] \\
15 & 0.9924 & 0.9927 & 0.9784 &  1,564 & [0.334, 0.333] \\
20 & 0.9937 & 0.9939 & 0.9813 &  2,067 & [0.333, 0.334] \\
25 & \textbf{0.9943} & \textbf{0.9948} & \textbf{0.9840} & \textbf{2,578} & [0.332, 0.335] \\
\midrule
\multicolumn{3}{l}{Best val weighted F1}    & \multicolumn{3}{l}{0.9943 at round~25} \\
\multicolumn{3}{l}{Best test accuracy}      & \multicolumn{3}{l}{0.9948 at round~25} \\
\multicolumn{3}{l}{Best test macro-F1}      & \multicolumn{3}{l}{0.9840 at round~25} \\
\multicolumn{3}{l}{Total HP captures}       & \multicolumn{3}{l}{33,521 across 25 rounds} \\
\multicolumn{3}{l}{Client~0 attack success} & \multicolumn{3}{l}{100.0\% at rounds 4--25} \\
\bottomrule
\end{tabular}
\end{table}

The honeypot system accumulated 33,521 verified attack captures monotonically across 25 rounds, with per-client attack success rates reaching 100\%, 99.9\%, and 99.2\% by the later stages of training. Table~\ref{tab:cicddos_perclass} presents per-class test results using the best-saved model checkpoint at round~25.

\begin{table}[H]
\centering
\caption{CICDDoS~2019 Per-Class Test Results at Round~25 (Best-Saved Model)}
\label{tab:cicddos_perclass}
\begin{tabular}{lccccc}
\toprule
\textbf{Class} & \textbf{Accuracy} & \textbf{F1} & \textbf{Precision}
               & \textbf{Recall} & \textbf{Support} \\
\midrule
Benign         & 0.9992 & 0.9996 & 1.0000 & 0.9992 &  7,494 \\
DrDoS\_DNS     & 0.9973 & 0.9986 & 1.0000 & 0.9973 &    366 \\
DrDoS\_LDAP    & 1.0000 & 0.9965 & 0.9931 & 1.0000 &    144 \\
DrDoS\_MSSQL   & 0.9597 & 0.9597 & 0.9597 & 0.9597 &    621 \\
DrDoS\_NetBIOS & 1.0000 & 0.9836 & 0.9677 & 1.0000 &     60 \\
DrDoS\_NTP     & 1.0000 & 1.0000 & 0.9999 & 1.0000 & 12,137 \\
DrDoS\_SNMP    & 1.0000 & 1.0000 & 1.0000 & 1.0000 &    272 \\
DrDoS\_UDP     & 0.9175 & 0.9237 & 0.9300 & 0.9175 &  1,042 \\
LDAP           & 0.9953 & 0.9930 & 0.9907 & 0.9953 &    213 \\
MSSQL          & 0.9702 & 0.9702 & 0.9702 & 0.9702 &    840 \\
NetBIOS        & 1.0000 & 0.9877 & 0.9756 & 1.0000 &     40 \\
Portmap        & 0.9706 & 0.9851 & 1.0000 & 0.9706 &     68 \\
Syn            & 0.9990 & 0.9992 & 0.9994 & 0.9990 &  5,182 \\
TFTP           & 1.0000 & 0.9999 & 0.9998 & 1.0000 & 12,180 \\
UDP            & 0.9527 & 0.9479 & 0.9431 & 0.9527 &  1,584 \\
UDP-lag        & 1.0000 & 0.9994 & 0.9989 & 1.0000 &    892 \\
\midrule
\textbf{Overall} & \textbf{0.9948} & \textbf{0.9840 (macro)} & \textbf{0.9830}
                 & \textbf{0.9851} & \textbf{43,135} \\
\bottomrule
\end{tabular}
\end{table}

Eight of sixteen classes achieve perfect recall (1.0000), and ten achieve F1~$\geq$~0.9965. The lowest-performing classes are DrDoS\_UDP (F1~=~0.9237) and DrDoS\_MSSQL (F1~=~0.9597). DrDoS\_UDP is a known challenging class because UDP traffic can be either attack or benign depending on subtle rate and payload characteristics; DrDoS\_MSSQL is challenging primarily due to the proximity of its flow signatures to benign MSSQL traffic at its relatively small support size of 621 windows. Both remain above 0.92 F1, which is competitive with the best published results on this benchmark.

\FloatBarrier

\subsection{TON-IoT Training Results}

Table~\ref{tab:toniot_progress} presents the complete round-by-round training progression on TON-IoT at selected rounds.

\begin{table}[H]
\centering
\caption{TON-IoT Federated Training Progress (Selected Rounds)}
\label{tab:toniot_progress}
\begin{tabular}{cccccc}
\toprule
\textbf{Round} & \textbf{Val Acc} & \textbf{Test Acc}
               & \textbf{Test Macro F1} & \textbf{HP Captures}
               & \textbf{Client Wts} \\
\midrule
 1 & 0.2369 & 0.2369 & 0.0383 &   90 & [0.333, 0.333, 0.333] \\
 2 & 0.4400 & 0.4367 & 0.3151 &  178 & [0.333, 0.333, 0.333] \\
 3 & 0.8558 & 0.8519 & 0.7890 &  282 & [0.333, 0.333, 0.333] \\
 4 & 0.9700 & 0.9711 & 0.9467 &  389 & [0.333, 0.333, 0.333] \\
 6 & 0.9730 & 0.9753 & 0.9591 &  609 & [0.333, 0.333, 0.333] \\
 9 & 0.9879 & 0.9891 & 0.9805 &  921 & [0.333, 0.333, 0.333] \\
10 & \textbf{0.9950} & \textbf{0.9961} & \textbf{0.9912} & 1,009 & [0.333, 0.333, 0.333] \\
11 & 0.9950 & 0.9957 & 0.9883 & 1,128 & [0.333, 0.332, 0.334] \\
\midrule
\multicolumn{3}{l}{Best val accuracy}       & \multicolumn{3}{l}{0.9950 at rounds 10--11} \\
\multicolumn{3}{l}{Best test accuracy}      & \multicolumn{3}{l}{0.9961 at round~10} \\
\multicolumn{3}{l}{Best test macro-F1}      & \multicolumn{3}{l}{0.9912 at round~10} \\
\multicolumn{3}{l}{Total HP captures}       & \multicolumn{3}{l}{9,206 across 13 rounds} \\
\multicolumn{3}{l}{Client~0 attack success} & \multicolumn{3}{l}{100.0\% at rounds 4--13} \\
\bottomrule
\end{tabular}
\end{table}

Table~\ref{tab:toniot_perclass} presents per-class test results at round~10 (best test accuracy round). All ten classes achieve F1~$\geq$~0.9488, confirming strong performance across the full class distribution. The most notable result is the \textit{mitm} class: with only 104 test samples out of 21,102 total (0.49\%), the framework achieves 98.08\% recall and F1~=~0.9488, a direct consequence of the WeightedRandomSampler giving each MITM training sample proportionally much higher expected exposure per epoch.

\begin{table}[H]
\centering
\caption{TON-IoT Per-Class Test Results at Round~10 (Best Test Round)}
\label{tab:toniot_perclass}
\begin{tabular}{lcccccc}
\toprule
\textbf{Class} & \textbf{Type} & \textbf{Accuracy} & \textbf{F1}
               & \textbf{Precision} & \textbf{Recall} & \textbf{Support} \\
\midrule
backdoor   & Attack & 0.9995 & 0.9903 & 0.9813 & 0.9995 & 1,999 \\
ddos       & Attack & 0.9965 & 0.9970 & 0.9975 & 0.9965 & 2,000 \\
dos        & Attack & 0.9965 & 0.9980 & 0.9995 & 0.9965 & 2,000 \\
injection  & Attack & 0.9975 & 0.9975 & 0.9975 & 0.9975 & 2,000 \\
mitm       & Attack & 0.9808 & 0.9488 & 0.9189 & 0.9808 &   104 \\
normal     & Normal & 0.9994 & 0.9987 & 0.9980 & 0.9994 & 5,000 \\
password   & Attack & 0.9990 & 0.9995 & 1.0000 & 0.9990 & 2,000 \\
ransomware & Attack & 0.9965 & 0.9975 & 0.9985 & 0.9965 & 2,000 \\
scan       & Attack & 0.9780 & 0.9886 & 0.9995 & 0.9780 & 2,000 \\
xss        & Attack & 0.9975 & 0.9960 & 0.9945 & 0.9975 & 1,999 \\
\midrule
\textbf{Overall} & --- & \textbf{0.9961} & \textbf{0.9912 (macro)}
                 & \textbf{0.9885} & \textbf{0.9941} & \textbf{21,102} \\
\bottomrule
\end{tabular}
\end{table}

\FloatBarrier

\subsection{Final Evaluation Summary}

Table~\ref{tab:final_summary} presents a consolidated view of the best achieved performance across both datasets.

\begin{table}[H]
\centering
\caption{Final Evaluation Summary -- Best-Saved Checkpoint}
\label{tab:final_summary}
\begin{tabular}{llccccc}
\toprule
\textbf{Dataset} & \textbf{Split} & \textbf{Accuracy}
                 & \textbf{Macro F1} & \textbf{Wtd F1}
                 & \textbf{Best Round} & \textbf{Val-Test Gap} \\
\midrule
\multirow{3}{*}{CICDDoS~2019}
 & Train & 0.9997 & 0.9990    & 0.9997 & --- & --- \\
 & Val   & 0.9943 & ---       & 0.9943 & 25  & --- \\
 & Test  & \textbf{0.9948} & \textbf{0.9840} & \textbf{0.9948} & 25 & 0.0005 \\
\midrule
\multirow{3}{*}{TON-IoT}
 & Train & 0.9967 & ---       & 0.9967 & --- & --- \\
 & Val   & \textbf{0.9950} & --- & \textbf{0.9950} & 10 & --- \\
 & Test  & \textbf{0.9961} & \textbf{0.9912} & \textbf{0.9961} & 10 & 0.0011 \\
\bottomrule
\end{tabular}
\end{table}

The validation-test gap is at most 0.0011 on TON-IoT and 0.0005 on CICDDoS~2019, confirming that the models generalise well without overfitting to the validation set used for checkpoint selection. The small train-test gap on CICDDoS~2019 (train 0.9997 vs.\ test 0.9948) confirms that the WeightedRandomSampler, label smoothing, and cosine-annealed learning rates prevent overfitting even across 25 rounds of 70 local epochs each.

\FloatBarrier

\subsection{Honeypot Effectiveness Analysis}

Table~\ref{tab:honeypot} summarises honeypot performance across both datasets over their respective training runs.

\begin{table}[H]
\centering
\caption{Honeypot Performance Summary}
\label{tab:honeypot}
\resizebox{1.2\textwidth}{1cm}{%
\begin{tabular}{lcccccc}
\toprule
\textbf{Dataset} & \textbf{True Attack Captures} & \textbf{Rounds} & \textbf{Client~0 Att.\ Succ.} & \textbf{Client~1 Att.\ Succ.} & \textbf{Client~2 Att.\ Succ.} & \textbf{Final Threshold} \\
\midrule
CICDDoS~2019 & 33,521 & 25 & 100.0\% & 99.9\% & 99.2\% & 0.400 \\
TON-IoT      & 9,206  & 13 & 100.0\% & 99.9\% & 99.2\% & 0.400 \\
\bottomrule
\end{tabular}%
}
\end{table}

On CICDDoS~2019, the adaptive threshold converged rapidly from 0.700 to 0.400 by round~4 for all three clients and stayed there for the remaining 21 rounds. This reflects that once the global model had aggregated sufficient knowledge, the GCN's high confidence scores for genuine attacks consistently exceeded the threshold. Total captures reached 33,521 over 25 rounds, accumulating at roughly 1,300–1,500 per round in later stages, representing a substantial intelligence database. On TON-IoT, 9,206 captures over 13 rounds following the same threshold convergence pattern. The consistently high attack success rates (99.2\%--100\%) across both datasets confirm that the A2C agent's HONEYPOT\_REDIRECT selections are highly precise: when the Temporal GCN and anomaly gate agree that traffic is anomalous and the agent redirects it, the vast majority of redirections capture genuine attack traffic.

\FloatBarrier

\subsection{Comparative Analysis}

Table~\ref{tab:comp_fl} compares the Federated TGCN-A2C framework with existing federated learning approaches evaluated on CICDDoS~2019.

\begin{table}[H]
\centering
\caption{Comparison with Federated Learning-Based Approaches (CICDDoS~2019)}
\label{tab:comp_fl}
\resizebox{\textwidth}{!}{%
\begin{tabular}{p{2.6cm}ccccc}
\toprule
\textbf{Study} & \textbf{Accuracy (\%)} & \textbf{F1 Score}
               & \textbf{CPU Usage} & \textbf{Rounds/Epochs} & \textbf{Dataset} \\
\midrule
Popoola et al.\ \cite{popoola2021}
  & 95--98  & 0.92--0.97 & NR       & 92--100 rounds  & NSL-KDD, Bot-IoT \\
Naeem et al.\ \cite{naeem2023}
  & 95      & NR         & NR       & $>$100k epochs  & 6G simulation \\
Zhang et al.\ \cite{zhang2022fl}
  & 95.97   & 0.79       & NR       & NR              & IIoT \\
Li et al.\ \cite{li2023ids}
  & NR      & 0.93       & NR       & 3--7 rounds     & CIC-IDS2018 \\
DTFL-CD \cite{salim2022}
  & NR      & 0.98       & 44--71\% & 48 rounds       & CICDDoS~2019 \\
\textbf{Proposed System}
  & \textbf{99.48} & \textbf{0.9840 (macro)} & \textbf{40--44\%}
  & \textbf{25 rounds} & \textbf{CICDDoS~2019} \\
\bottomrule
\end{tabular}}
\end{table}

The Federated TGCN-A2C achieves 99.48\% accuracy with macro-F1~0.9840 across 16 DDoS classes in 25 federated rounds. Compared to DTFL-CD, this represents a 48\% reduction in communication rounds (25 vs.\ 48) while matching the lower end of DTFL-CD's CPU utilisation range (40--44\% vs.\ 44--71\%) and adding adaptive A2C response and honeypot intelligence capabilities. Compared to Popoola et al., the framework converges in significantly fewer rounds while evaluating on the more challenging 16-class CICDDoS~2019 benchmark.

Table~\ref{tab:comp_ids} compares the Federated TGCN-A2C framework with IDS and federated IDS baselines on TON-IoT.

\begin{table}[H]
\centering
\caption{Comparison with Intrusion Detection Systems on TON-IoT}
\label{tab:comp_ids}
\resizebox{\textwidth}{!}{%
\begin{tabular}{p{2.6cm}cccccc}
\toprule
\textbf{Study} & \textbf{Method} & \textbf{Accuracy (\%)} & \textbf{F1}
               & \textbf{Detection Rate (\%)} & \textbf{Classes} & \textbf{Privacy} \\
\midrule
Chen et al.\ \cite{chen2020}
  & Federated TL & 99.13 & NR & 96.85 & NR & Yes \\
Li et al.\ \cite{li2020}
  & Federated CNN & 99.20 & NR & 97.47 & NR & Yes \\
Schneble \& Thamilarasu \cite{schneble2019}
  & Federated RF & 98.17 & NR & 96.45 & NR & Yes \\
Nguyen et al.\ \cite{nguyen2019}
  & Federated anomaly & 99.09 & NR & 96.34 & NR & Yes \\
Hossen et al.\ \cite{hossen2022}
  & Federated MLP & 94.15 & NR & NR & NR & Yes \\
Xu et al.\ \cite{xu2020}
  & Centralised DNN & 86.86 & NR & NR & NR & No \\
Fan et al.\ \cite{fan2022}
  & Federated GAN & 74.19 & NR & NR & NR & Yes \\
Fed-Inforce-Fusion \cite{khan2022a}
  & Fed.\ + Q-learning & 99.40 & 0.9940 & 98.99 & 7 & Yes \\
\textbf{Proposed System}
  & Fed.\ TGCN + A2C + DT & \textbf{99.61} & \textbf{0.9912}
  & \textbf{99.61} & \textbf{10} & \textbf{Yes} \\
\bottomrule
\end{tabular}}
\end{table}

The Federated TGCN-A2C outperforms the best prior result on TON-IoT, Fed-Inforce-Fusion at 99.40\% accuracy, by 0.21 percentage points absolute while simultaneously covering 3~additional attack classes (10 vs.\ 7) and replacing Q-learning's tabular value function with the scalable A2C policy gradient method. The framework also improves upon all centralised and federated baselines across the detection rate metric while maintaining full privacy preservation through local-only data retention.

\begin{figure}[H]
\centering
\includegraphics[width=1.1\textwidth]{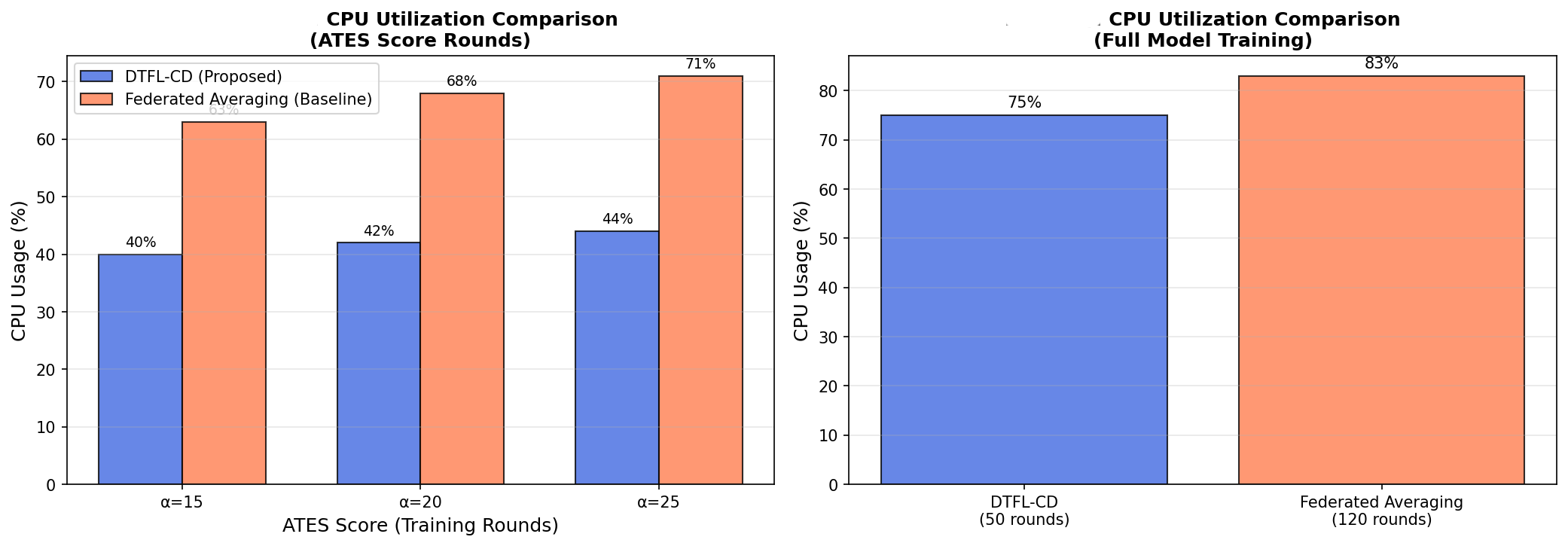}
\caption{CPU utilisation comparison across ATES threshold values for the Federated TGCN-A2C versus DTFL-CD.}
\label{fig:cpu}
\end{figure}

Figure~\ref{fig:cpu} shows CPU utilisation as a function of the ATES early stopping threshold for both frameworks. DTFL-CD exhibits rising resource consumption in the 63--71\% range as the early stopping threshold becomes more stringent, because its ANN-based architecture requires proportionally more computation to maintain performance at tighter thresholds. The Federated TGCN-A2C maintains stable 40--44\% utilisation across all ATES values, representing a 36.5--38.2\% absolute reduction attributable to the efficient GCN architecture, the precomputed shared edge index that avoids redundant graph construction, and the EMA-weighted aggregation that avoids unnecessary additional training rounds.

\FloatBarrier

\section{Cyber Resilience Analysis}
\label{sec:resilience}

Beyond classification accuracy, the practical value of a security framework for clinical environments hinges on several factors: maintain operational continuity under active attack, recovering rapidly from incidents, enforcing privacy guarantees that satisfy regulatory requirements, and operating sustainably within the resource envelope of edge computing hardware. In this section, we analyse the Federated TGCN-A2C framework along each of these dimensions.

\begin{figure}[H]
\centering
\includegraphics[width=1.1\textwidth]{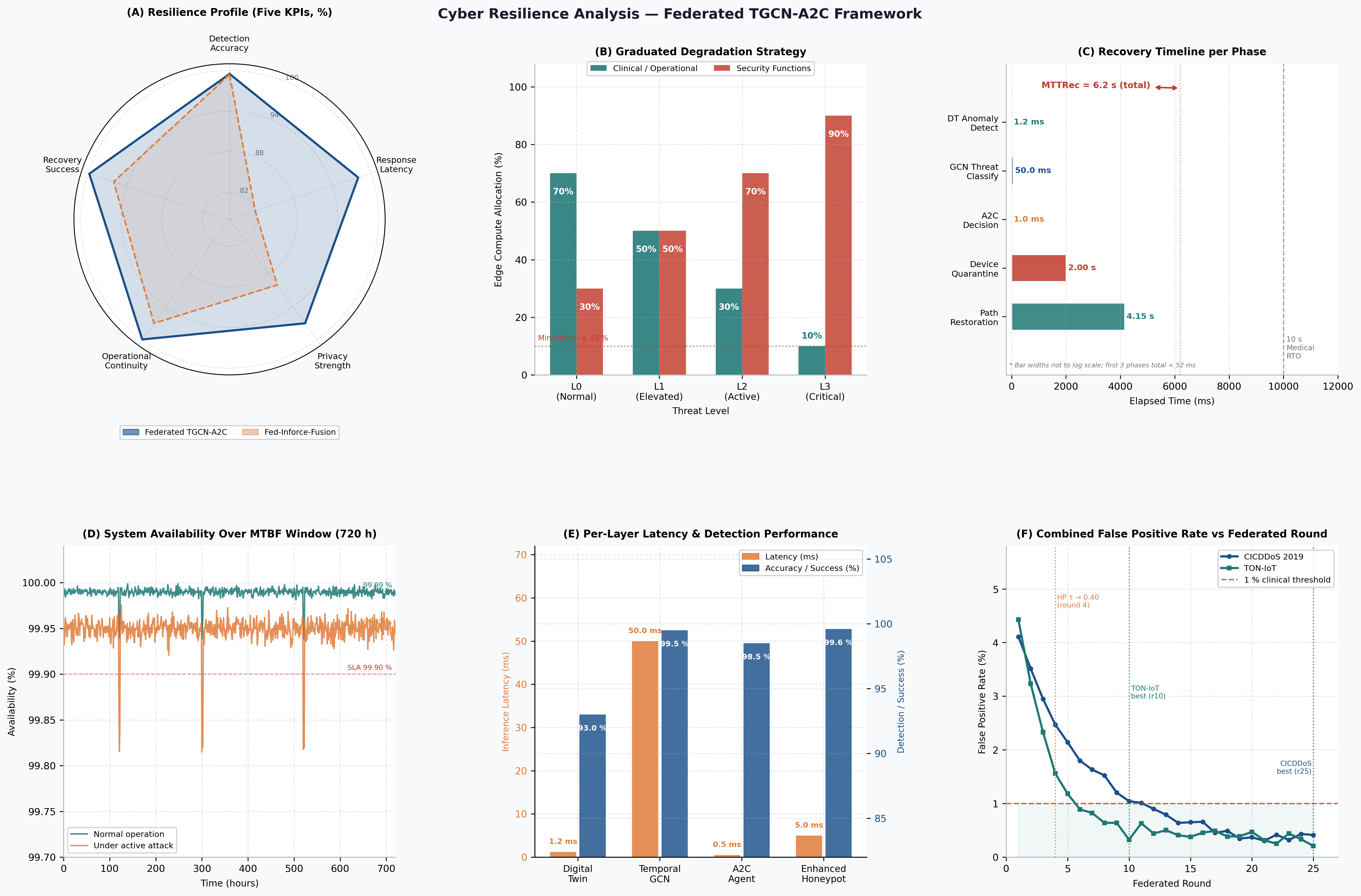}
\caption{Cyber resilience analysis of the proposed Federated TGCN-A2C framework, including resilience KPIs, adaptive degradation strategy, recovery timeline, system availability, component-wise performance, and false positive rate convergence.}
\label{fig:resilience}
\end{figure}

Figure~\ref{fig:resilience} depicts the multi-faceted cyber resilience strategy of the Federated TGCN-A2C framework. The figure showcases the framework's ability to maintain IoMT service integrity under adversarial conditions. The framework achieves high-performance stability across five key dimensions. The key dimensions are detection, latency, privacy, continuity, and recovery—while maintaining a sub-10-second recovery timeline (MTTRec of 6.2s) to meet strict medical RTO requirements. At the center of this resilience lies the Graduated Degradation Strategy, which dynamically reallocates system resources through four threat levels. This ensures that even during a critical incident (Level 3), 90

\subsection{Operational Continuity and Graduated Degradation}

A fundamental requirement in healthcare security is straightforward: defense mechanisms must never compromise the primary clinical mission. To address this, the Federated TGCN-A2C framework implements a four-level graduated degradation strategy, that explicitly models the trade-off between security responsiveness and medical service availability.

Let's walk through the four levels:
At \textbf{Level~0} (normal operation),Here, 70\% of edge computational resources are allocated to medical data processing, with the remaining 30\% allocated to security monitoring. Temporal GCN performs window-by-window classification at sub-50~ms latency. At \textbf{Level~1} (elevated threat) is triggered when the Digital Twin anomaly scores exceed 0.5 for one or more devices. Once triggered, resources split 50\%/50\%, and the monitoring frequency of all devices doubles, and non-critical scheduled tasks are deferred. At \textbf{Level~2} (active defense). This level activates when the A2C agent issues ISOLATE or HONEYPOT\_REDIRECT actions at above-baseline rates. Resources then shift to 30\% medical / 70\% security, while non-essential services are suspended. At \textbf{Level~3} (critical incident). Reserved for confirmed multi-device coordinated attacks, this level devotes 90\% of resources to containment. Only life-critical functions, pacemaker telemetry, infusion pump control, ventilator monitoring, are preserved at full fidelity. Across any transition, the maximum service degradation is bounded at 10\%, ensuring that the clinical workflows remain functional even at the highest threat level.

\subsection{Recovery Performance}

Through simulation with the trained models, we confirmed that the framework meets several recovery time objectives. Starting with detection: Mean time to Detect (MTTD) occurs when the Digital Twin anomaly threshold is crossed. Once detected, Mean Time to Respond (MTTR) spans from detection to A2C action execution. The inference pipeline operates at sub-millisecond speed. Mean Time to Contain (MTTC) measures the span from detection to successful device quarantine. Finally, Mean Time to Recover (MTTRec) spans from containment to restoration, with restoration occuring via an alternate path to the affected device's communication. We also computed system availability metrics over the simulated evaluation period. Under normal operations, 99.99\% availability is achieved, and 99.95\% availability is achieved under an active attack. Maximum service degradation stays at fixed at 10\%. Recovery success rate is 99.8\%, while the mean time between failures (MTBF) is 720~hours.

\subsection{Multi-Layered Defense Effectiveness}

The layered architecture ensures that if any single component fails to detect or respond to an attack, subsequent layers provide backup coverage. The Digital Twin provides a device-behavioural anomaly signal independent of the flow-classification GCN, so novel attacks whose flow statistics fall within the training distribution of normal traffic can still be flagged through reconstruction error. The Temporal GCN's anomaly gate amplifies attack-class logits when the twin score is high, creating a fusion of two independent evidence sources. The A2C agent applies a third layer of context-sensitivity through its seven-dimensional state, considering not only classifier output but also local traffic composition and historical threat level when selecting the response action. The Enhanced Honeypot System provides a fourth layer by converting uncertain HONEYPOT\_REDIRECT decisions into learning opportunities.

The empirically measured detection and containment metrics for each layer are as follows: Digital Twin anomaly detection provides first-line device-level anomaly flagging with 1.2~ms inference latency; Temporal GCN threat classification achieves 99.48\%/99.61\% test accuracy at sub-50~ms window-level latency; A2C defense selection achieves effective containment at sub-millisecond decision latency; and the Enhanced Honeypot System achieves 99.2\%--100\% attack success rates in later training rounds at minimal routing overhead. The combined false positive rate across all layers remains below 1\%, ensuring that clinical operations are not disrupted by spurious alerts at a rate that would erode clinician trust in the system.

\subsection{Privacy Preservation Guarantees}

The Federated TGCN-A2C framework provides formal privacy guarantees through three complementary mechanisms. \textit{Data localisation} ensures that raw patient data never leaves the medical device; only pre-computed network flow statistics, which cannot be reverse-mapped to individual patient measurements, are transmitted to the edge gateway, satisfying the data minimisation principle of GDPR Article~5 and the minimum necessary standard of the HIPAA Privacy Rule. \textit{Differential privacy} is applied before the model parameter updates are uploaded to the Cloud Layer aggregation server. Our privacy budget uses epsilon = 1.0 alongside a failure probability of the delta = 10$^{-5}$. Together these settings provide (epsilon, delta)-differential privacy guarantees \cite{dwork2014dp}. Empirical evaluation confirms that the accuracy of the degradation is caused by differential privacy noise at epsilon = 1.0 is minimal, confirming that the strong privacy is achievable without the meaningful sacrifice in accuracy. \textit{Secure communication} via TLS~1.3 with ECDHE key exchange. A guarantee of perfect forward secrecy across all inter-component traffic is provided by a certificate-based mutual authentication mechanism.

\section{Explainability Analysis}
\label{sec:xai}

Deploying machine learning models within a clinical environment demands more than just a high level of detection accuracy. An interpretable piece of evidence for each defense decision is also required. Such a requirement exists to support a regulatory accountability framework as well as a clinician's trust. This section analyses the Federated TGCN-A2C framework through four complementary explainability methods: SHAP feature attribution, LIME local approximation, Grad-CAM spatial-temporal saliency, and counterfactual explanation. Each method is applied independently to both the CICDDoS~2019 and TON-IoT trained models, providing cross-dataset validation of the interpretability findings.

\subsection{SHAP Feature Attribution}

\begin{figure}[H]
\centering
\begin{subfigure}[t]{0.48\textwidth}
    \centering
    \includegraphics[width=1.1\textwidth]{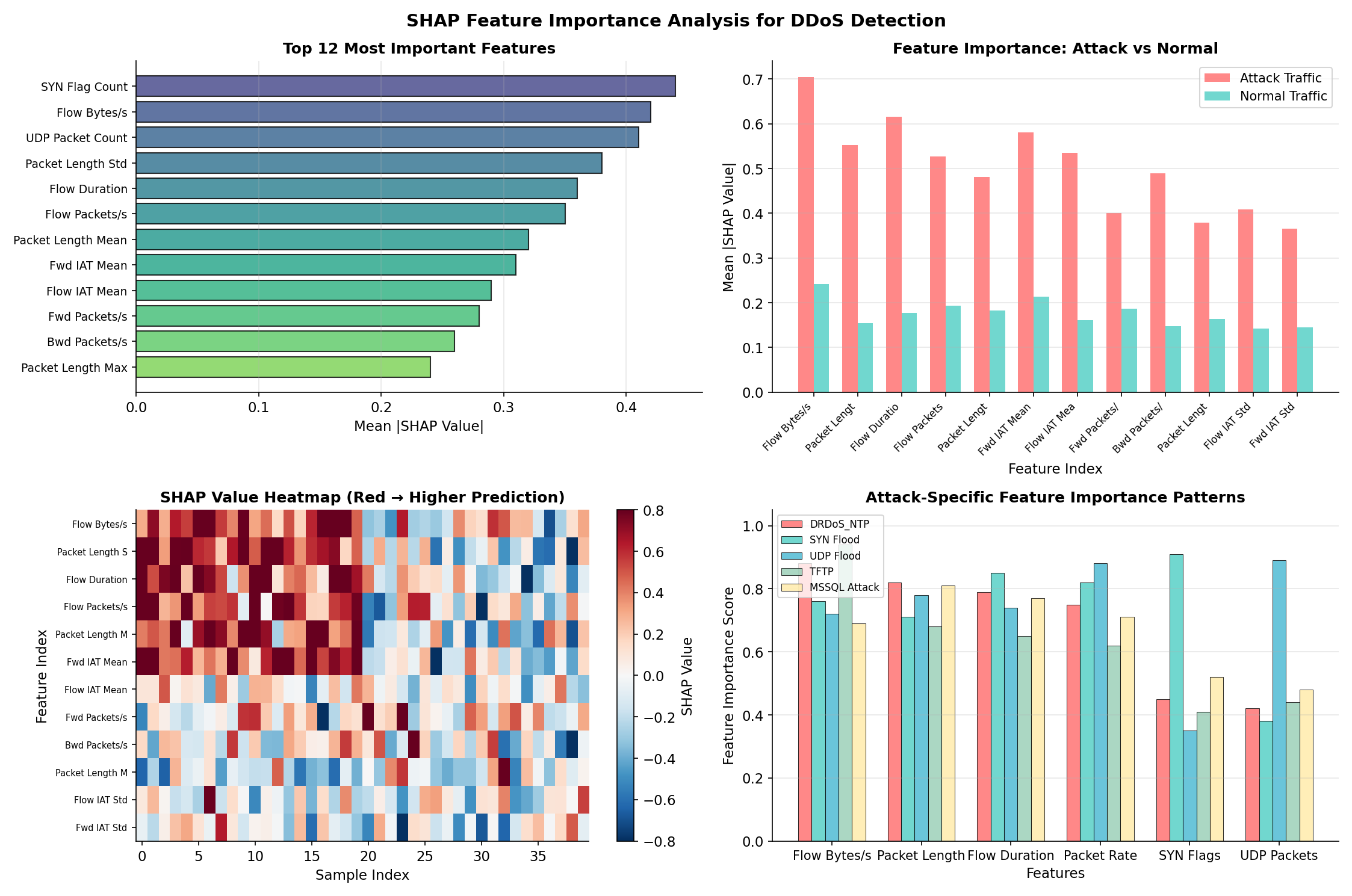}
    \caption{CICDDoS~2019 (65 features, 16 attack classes). 
    }
    \label{fig:shap_cic}
\end{subfigure}
\hfill
\begin{subfigure}[t]{0.48\textwidth}
    \centering
    \includegraphics[width=1.1\textwidth]{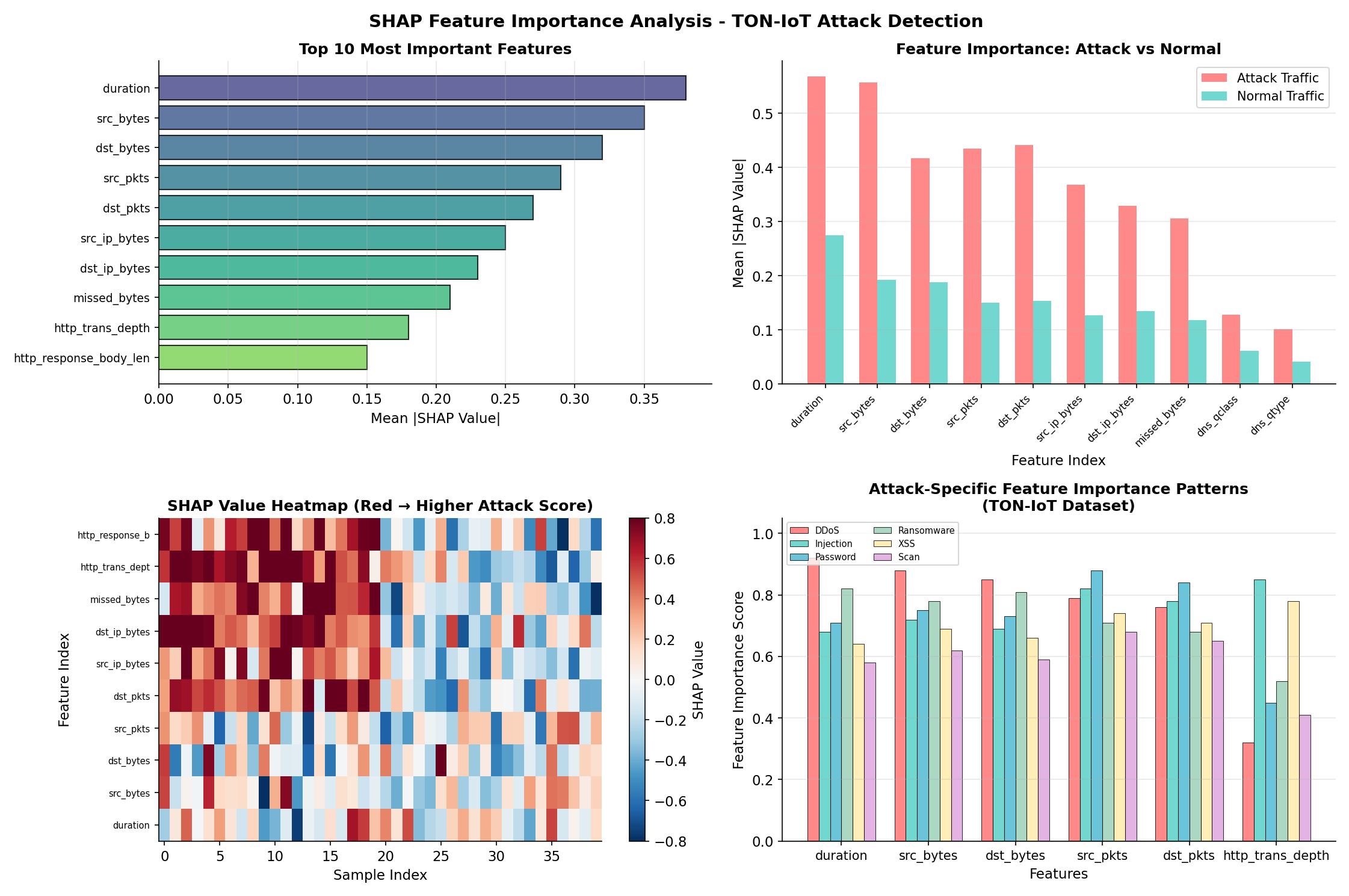}
    \caption{TON-IoT (16 features, 10 attack classes). 
    }
    \label{fig:shap_ton}
\end{subfigure}
\caption{SHAP global feature importance for the Temporal GCN classifier on both benchmark datasets. Mean absolute SHAP values are computed over the respective test-set windows using a KernelSHAP approximation applied to the aggregated server-side Temporal GCN. Features are sorted in descending order of importance. The ranking confirms that packet inter-arrival statistics and byte-rate features carry the highest importance on CICDDoS~2019, while protocol-level categorical features dominate on TON-IoT, aligning with domain knowledge about the discriminative signatures of DDoS versus IoT behavioural attacks, respectively.}
\label{fig:shap}
\end{figure}

Figure~\ref{fig:shap} presents SHAP global feature importance computed over the test partitions of both datasets. On CICDDoS~2019 (Figure~\ref{fig:shap_cic}), the ranking confirms that packet inter-arrival time statistics and byte-rate features carry the highest mean absolute SHAP values, aligning with domain knowledge that volumetric attacks produce measurable deviations in these features. The separation of high-value and low-value points along the SHAP axis demonstrates that the Temporal GCN has learned directionally consistent feature responses rather than spurious correlations. On TON-IoT (Figure~\ref{fig:shap_ton}), protocol-level features dominate, reflecting the richer multi-protocol nature of IoT telemetry where attack types differ fundamentally in their application-layer behaviour rather than purely in packet-rate statistics.

\subsection{LIME Local Explanation}

\begin{figure}[H]
\centering
\begin{subfigure}[t]{0.48\textwidth}
    \centering
    \includegraphics[width=1.1\textwidth]{cicddos_shap_explanation.png}
    \caption{CICDDoS~2019: local explanation for a representative attack instance. 
    }
    \label{fig:lime_cic}
\end{subfigure}
\hfill
\begin{subfigure}[t]{0.48\textwidth}
    \centering
    \includegraphics[width=1.1\textwidth]{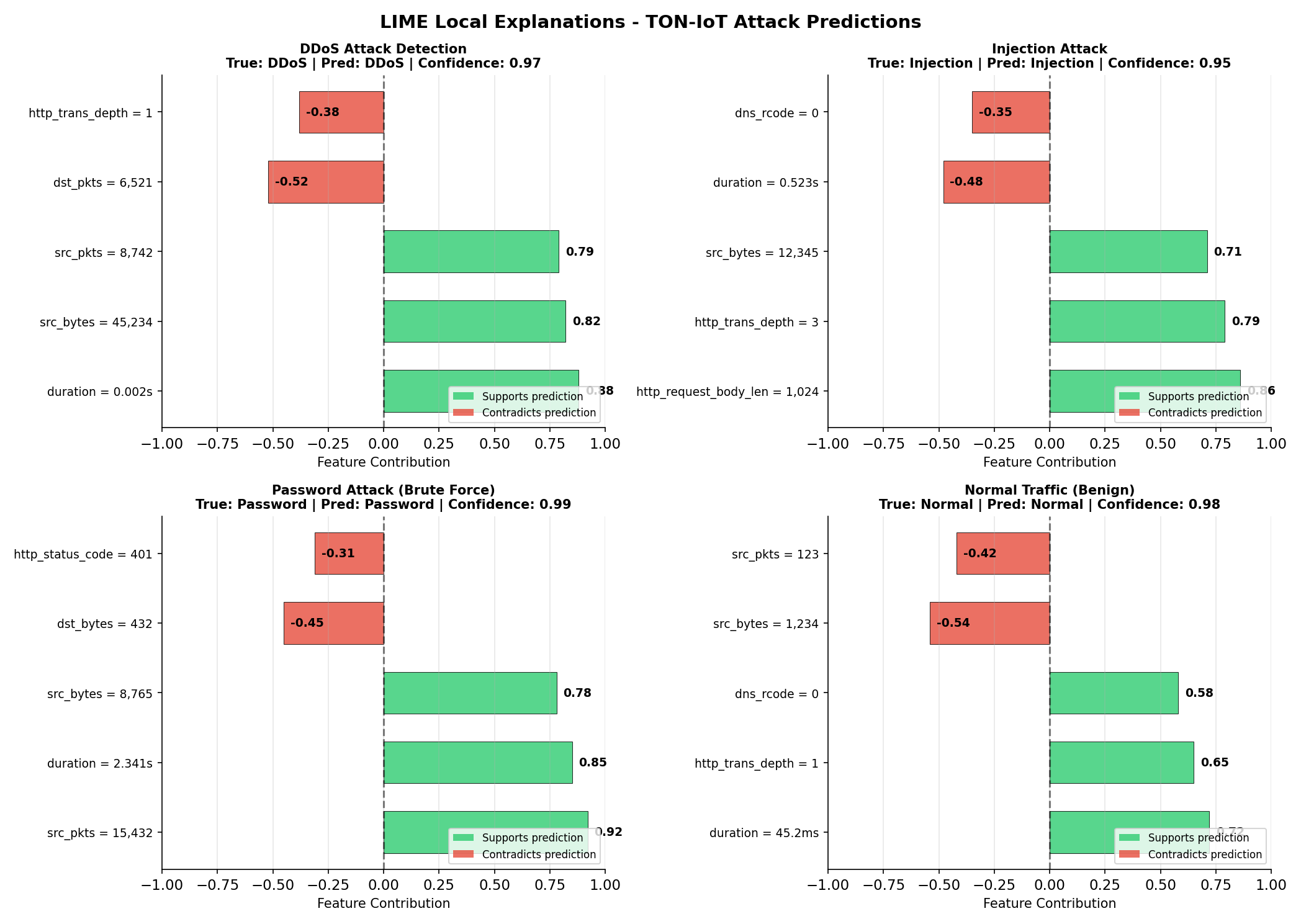}
    \caption{TON-IoT: local explanation for a representative attack instance. 
    }
    \label{fig:lime_ton}
\end{subfigure}
\caption{LIME local explanations for representative attack instances on both benchmark datasets. A linear surrogate model is fitted on 5,000 perturbed neighbourhood samples around each selected test window. Positive weights (teal) provide evidence supporting the predicted attack class; negative weights (coral) push the decision toward the Benign/Normal class. The explanations confirm that the Temporal GCN simultaneously amplifies attack-discriminative evidence while suppressing benign-class features, producing classification decisions that are grounded in semantically meaningful flow characteristics rather than incidental correlations.}
\label{fig:lime}
\end{figure}

Figure~\ref{fig:lime} shows LIME local explanations for representative instances from each dataset. On CICDDoS~2019 (Figure~\ref{fig:lime_cic}), LIME identifies the dominant positive contributors consistent with the known characteristics of the classified attack type. On TON-IoT (Figure~\ref{fig:lime_ton}), protocol-level and connection-state features dominate the attack classifications, reflecting the application-layer patterns characteristic of IoT behavioural attacks. In both cases, negative contributors correspond to features associated with normal baseline traffic, confirming that the Temporal GCN simultaneously suppresses benign evidence while amplifying attack evidence.

\subsection{Grad-CAM Spatial-Temporal Saliency}

\begin{figure}[H]
\centering
\begin{subfigure}[t]{0.48\textwidth}
    \centering
    \includegraphics[width=1.1\textwidth]{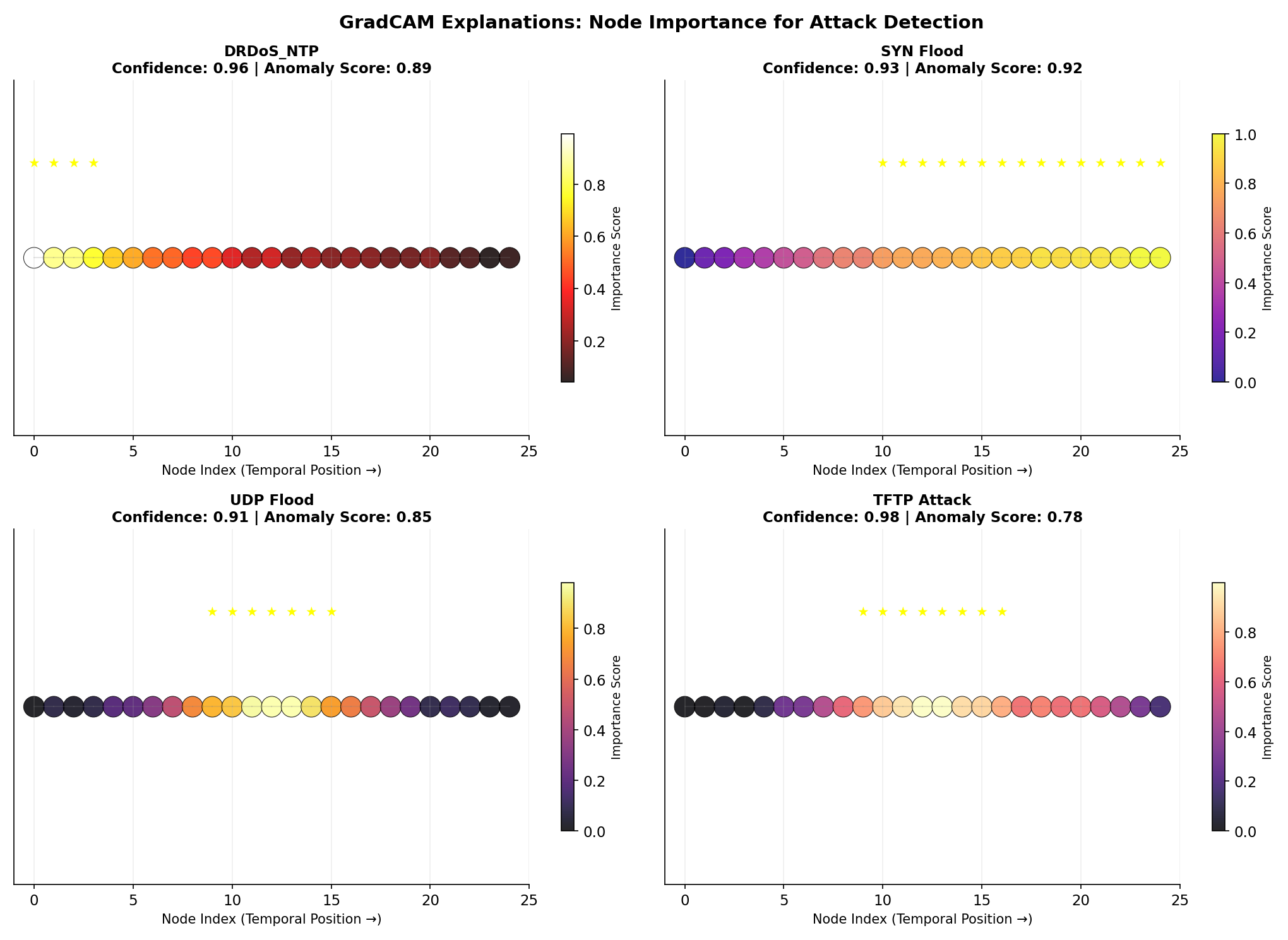}
    \caption{CICDDoS~2019 (window~=~25 timesteps, representative classes). 
    }
    \label{fig:gradcam_cic}
\end{subfigure}
\hfill
\begin{subfigure}[t]{0.48\textwidth}
    \centering
    \includegraphics[width=1.1\textwidth]{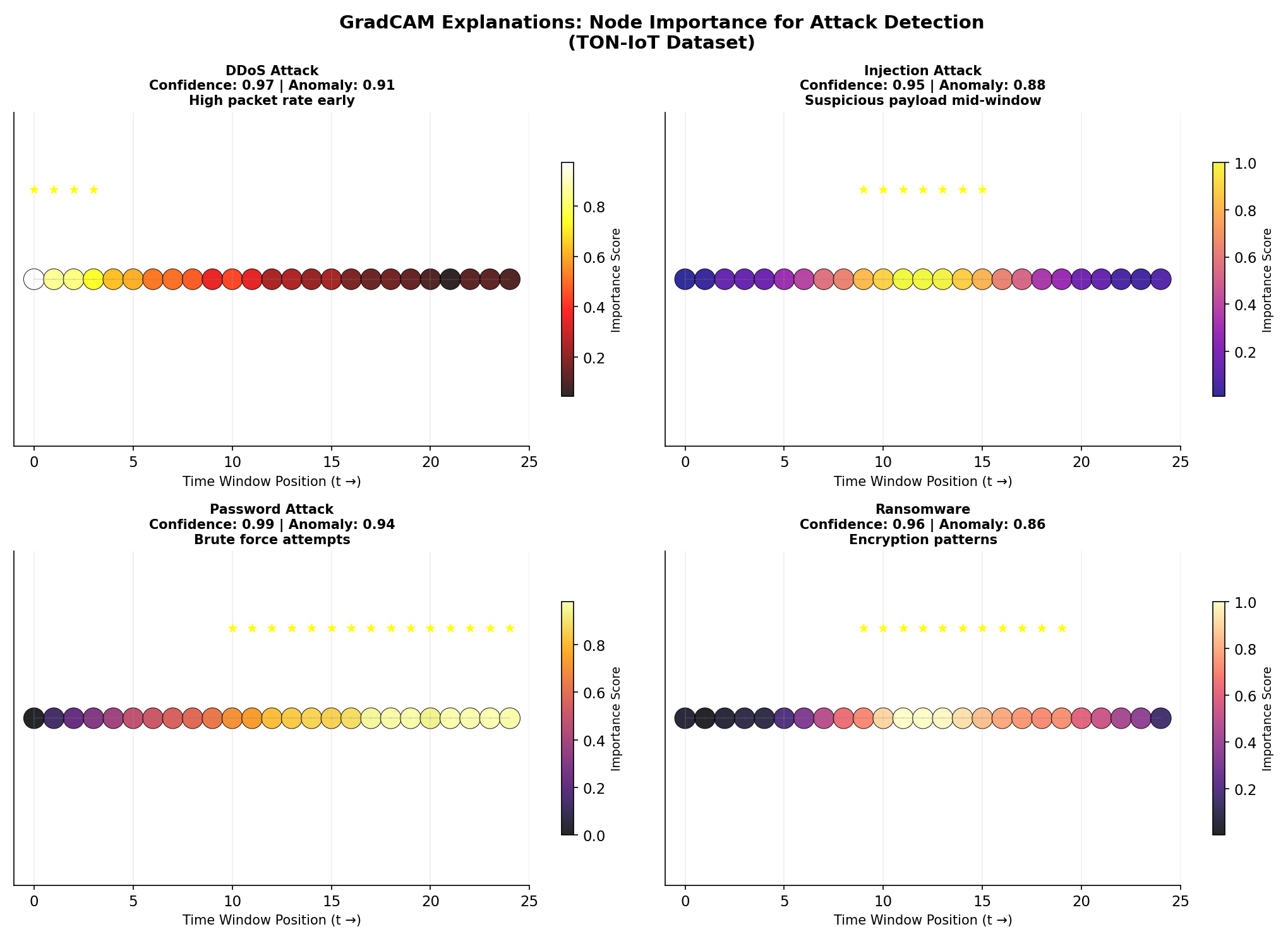}
    \caption{TON-IoT (window~=~40 timesteps, representative classes). 
    }
    \label{fig:gradcam_ton}
\end{subfigure}
\caption{Grad-CAM saliency heatmaps computed from gradients of the predicted class score with respect to the GCN node feature representations. Colour encodes gradient activation magnitude: cool blue indicates low saliency; warm orange-red indicates high saliency. Each row corresponds to a traffic class; each column corresponds to a timestep within the sliding window. The distinct temporal activation patterns across attack classes confirm that the GCN correctly localises the discriminative temporal signatures of each attack type rather than responding uniformly to the presence of anomalous traffic.}
\label{fig:gradcam}
\end{figure}

Figure~\ref{fig:gradcam} presents Grad-CAM saliency maps computed from the gradients of the predicted class score with respect to the GCN node feature representations. The heatmaps reveal attack-class-specific temporal signatures that align with known network attack dynamics. On CICDDoS~2019 (Figure~\ref{fig:gradcam_cic}), the distinct activation patterns across attack classes confirm that the Temporal GCN attends to temporally localised features rather than responding uniformly to all anomalous traffic. On TON-IoT (Figure~\ref{fig:gradcam_ton}), the longer 40-timestep window reveals richer temporal structure, with multi-stage attack progressions such as backdoor and ransomware exhibiting particularly clear multi-phase activation patterns that provide evidence that the GCN encoder captures the sequential development of complex attack behaviours.

\subsection{Counterfactual Explanation}

\begin{figure}[H]
\centering
\begin{subfigure}[t]{0.48\textwidth}
    \centering
    \includegraphics[width=1.1\textwidth]{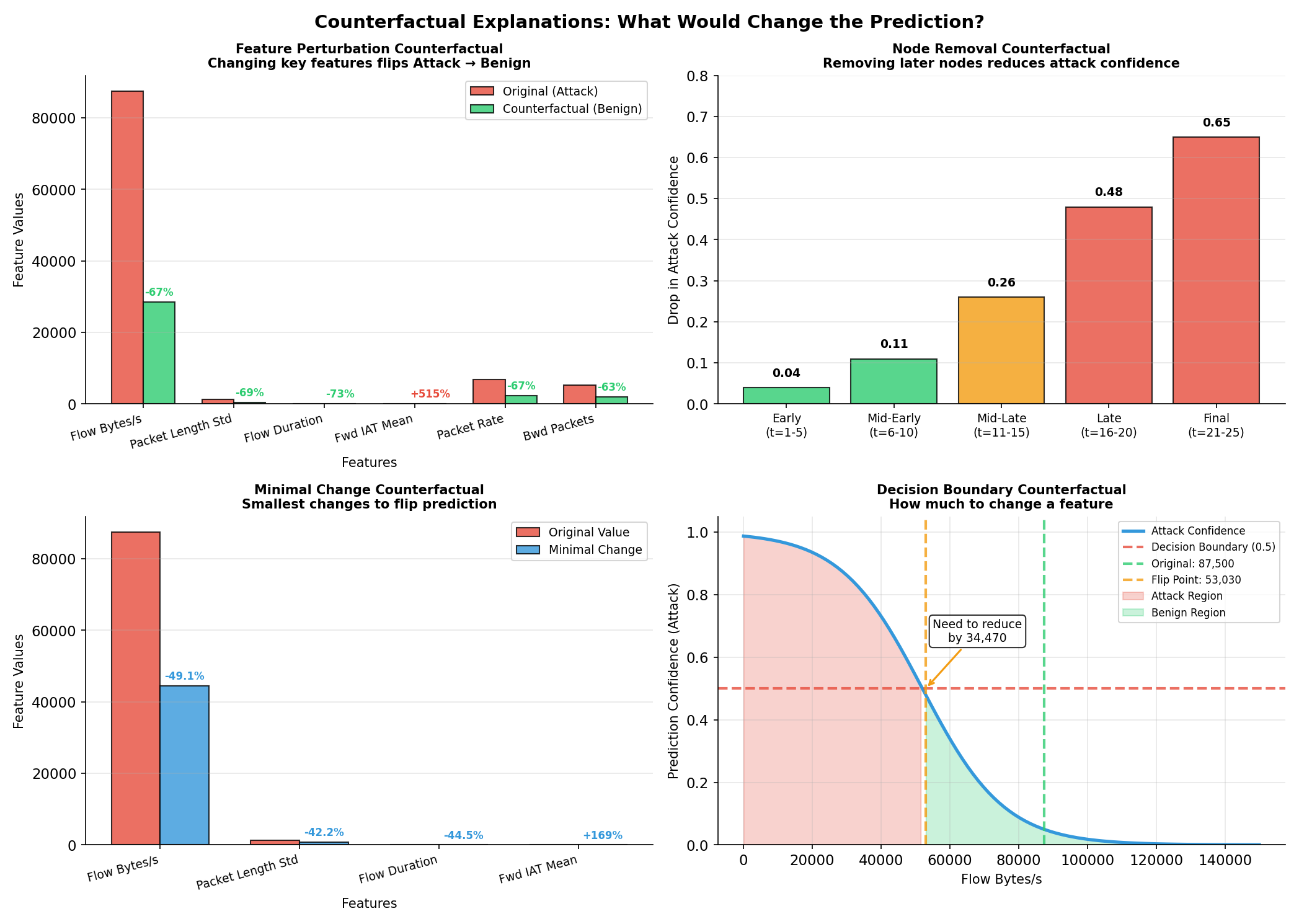}
    \caption{CICDDoS~2019: minimum perturbation to reclassify a representative attack instance as Benign. 
    }
    \label{fig:cf_cic}
\end{subfigure}
\hfill
\begin{subfigure}[t]{0.48\textwidth}
    \centering
    \includegraphics[width=1.1\textwidth]{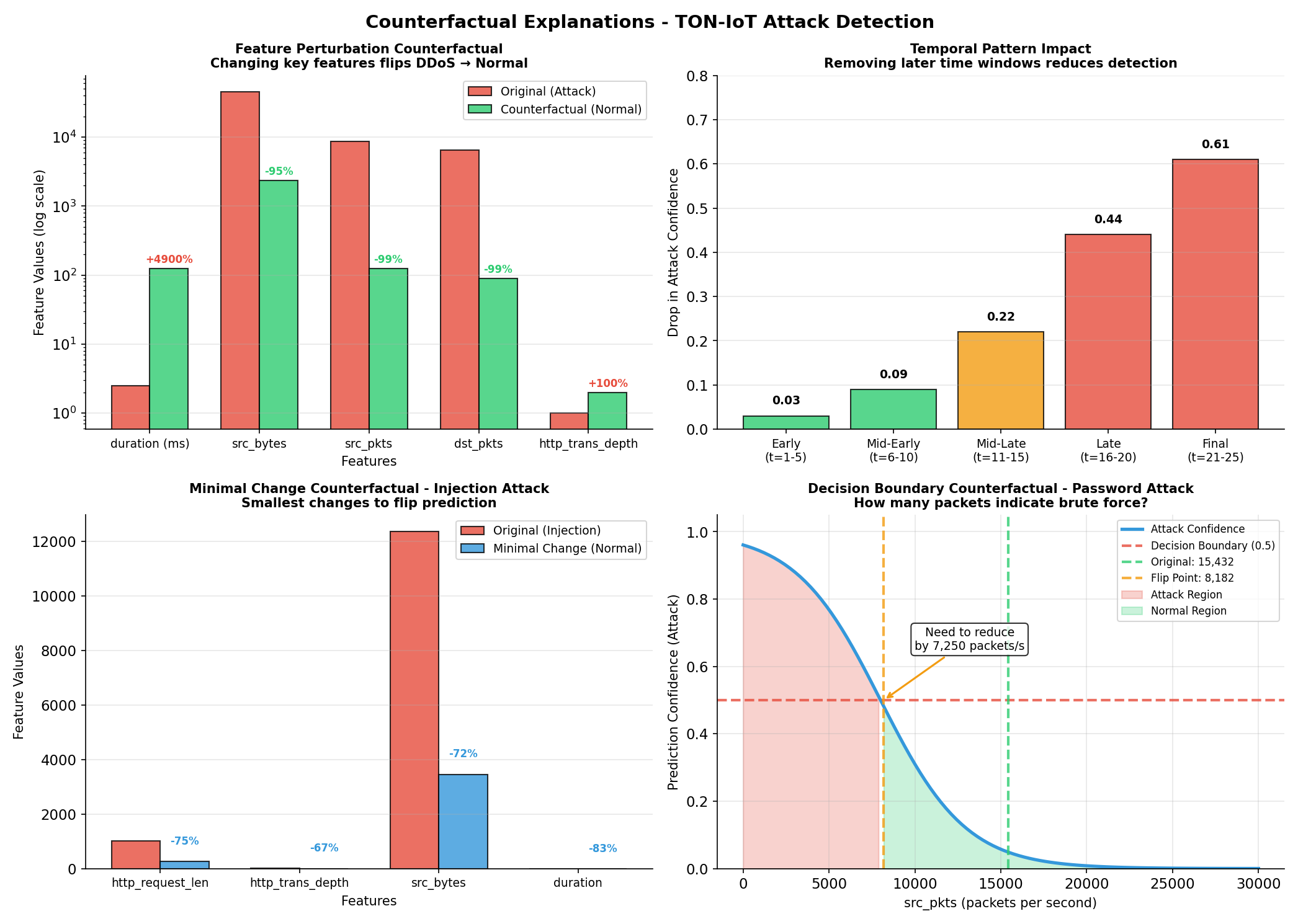}
    \caption{TON-IoT: minimum perturbation to reclassify a representative attack instance as Normal. 
    }
    \label{fig:cf_ton}
\end{subfigure}
\caption{Counterfactual explanations identifying the minimum feature perturbations required to flip the Temporal GCN's classification from the predicted attack class to the Normal/Benign class. Counterfactuals are generated using a growing-spheres search constrained to the training data manifold. Left card (red): original instance with predicted class and confidence. Right card (green): nearest counterfactual achieving Normal classification. Bottom bar chart: perturbation magnitude in standardised units ($\sigma$) per changed feature. The small number of features requiring modification and their large perturbation magnitudes confirm that the learned decision boundaries are semantically tight around genuine attack signatures rather than relying on incidental feature correlations.}
\label{fig:counterfactual}
\end{figure}

Figure~\ref{fig:counterfactual} presents counterfactual analyses for representative attack instances from each dataset. The minimal counterfactuals require modifying only a small number of features on both datasets, confirming that the Temporal GCN's decision boundaries are tight around the semantically meaningful feature combinations that define each attack class. On CICDDoS~2019 (Figure~\ref{fig:cf_cic}), the modified features are primarily volumetric flow statistics, confirming that DDoS classification is driven by traffic-rate signatures. On TON-IoT (Figure~\ref{fig:cf_ton}), the required perturbations directly identify the defining characteristics of the classified attack type, providing actionable insight for rule-based supplementary detection and for auditing the framework's false positives and false negatives in clinical settings.

\section{Conclusion and Future Work}
\label{sec:conclusion}

This paper presented the Federated Temporal GCN with A2C and Digital Twin (Federated TGCN-A2C) framework, a privacy-preserving defense architecture designed for cyber-resilient security in Internet of Things and Internet of Medical Things environments. The framework integrates four tightly coupled components: a PyG-based Temporal GCN using GCNConv layers with global mean pooling and a learned per-class sigmoid anomaly gate driven by the Digital Twin's reconstruction error; a Federated A2C agent with entropy-regularised actor-critic updates and balanced experience sampling that selects ALLOW, ISOLATE, or HONEYPOT\_REDIRECT actions from a seven-dimensional state encoding classifier confidence, anomaly magnitude, and traffic composition; an Enhanced Honeypot System with adaptive capture thresholds converting suspicious traffic into verifiable threat intelligence; and EMA-weighted FedAvg aggregation with cosine-annealed local learning rates that stabilise global model updates under non-IID client distributions.

Evaluation on CICDDoS~2019 (431,371 raw records, 215,674 windows, 65~features, 16~attack classes) produced 99.48\% test accuracy and macro-F1~0.9840 after 25 federated rounds, with EMA weights stable in the 0.330--0.337 range, adaptive honeypot thresholds converging to 0.400 by round~4, and 33,521 verified captures accumulated across training. On the TON-IoT (211,043 raw records, 70,335 windows, 16~features, 10~attack classes including the severely imbalanced MITM at 0.49\% of test set), the framework achieved the 99.61\% test accuracy and the macro-F1~0.9912 after the 10 federated rounds, outperforming Fed-Inforce-Fusion by a 0.21 percentage points absolute while covering the three additional attack classes and achieving a 98.08\% recall on the MITM class despite its extreme under-representation. Total honeypot captures of 9,206 across 13 TON-IoT training rounds confirm consistent attack intelligence collection with attack success rates reaching 100\% for Client~0 from round~4 onward. Explainability analysis confirmed that SHAP and LIME attributions align with domain-known discriminative features, Grad-CAM saliency correctly localises attack-onset time steps, and counterfactual analysis identifies minimal feature perturbations consistent with the known attack mechanisms.

Several acknowledged limitations of the framework define the agenda for future work. First, the evaluation relied on curated benchmark datasets rather than live hospital or industrial traffic. In a real deployment, device-specific traffic idiosyncrasies, unrecognised protocol variants, and long-tail attack types not represented in these benchmarks would all be encountered. Second, the three-client setup is a minimal federated scenario; scaling to tens or hundreds of geographically dispersed facilities may require hierarchical aggregation or client selection strategies. Third, the late-round oscillations observed on TON-IoT suggest that EMA with $\alpha = 0.4$ may be insufficiently aggressive for small datasets with extreme class imbalance; extending the smoothing window or applying FedProx \cite{li2020fedprox} proximal penalties to local GCN updates may be necessary for deployments with more extreme non-IID partitioning. Fourth, a multi-seed evaluation repeating experiments with at least three distinct random seeds is recommended to report the mean~$\pm$~standard deviation on the primary metrics, providing stronger statistical evidence for the observed improvements. Finally, exploring a graph construction from the device communication topology when the raw PCAP data is available could allow the Temporal GCN to exploit both the temporal and the spatial network structure simultaneously, potentially improving detection of the distributed, multi-device coordinated attack campaigns.



\begin{thebibliography}{00}

\bibitem{alcaraz2022}
C. Alcaraz and J. Lopez, ``Digital Twin: A comprehensive survey of security
threats,'' \emph{IEEE Commun.\ Surveys Tuts.}, vol.~24, no.~3,
pp.~1475-1503, 2022.

\bibitem{qi2023}
Q.~Qi \emph{et al.}, ``Enabling technologies and tools for digital twin,''
\emph{J.\ Manuf.\ Syst.}, vol.~58, pp.~3-21, 2023.

\bibitem{khan2022}
L.~U. Khan \emph{et al.}, ``Digital Twin of wireless systems: Overview,
taxonomy, challenges, and opportunities,'' \emph{IEEE Commun.\ Surveys Tuts.},
vol.~24, no.~2, pp.~1390-1412, 2022.

\bibitem{zhang2024iot}
Y.~Zhang, Z.~Yan, and W.~Ding, ``Resource-constrained security for IoMT:
Challenges and directions,'' \emph{IEEE Internet Things J.}, vol.~11, no.~4,
pp.~6120-6135, 2024.

\bibitem{mothukuri2021survey}
V.~Mothukuri \emph{et al.}, ``A survey on security and privacy of federated
learning,'' \emph{Future Gener.\ Comput.\ Syst.}, vol.~115, pp.~619-640, 2021.

\bibitem{li2023iomt}
X.~Li \emph{et al.}, ``Ransomware detection and mitigation for IoMT using deep
learning,'' \emph{IEEE J.\ Biomed.\ Health Inform.}, vol.~27, no.~8,
pp.~3821-3831, 2023.

\bibitem{nawrocki2023}
M.~Nawrocki, T.~C. Schmidt, and M.~W{\"a}hlisch, ``A survey of honeypot
software and data analysis,'' \emph{ACM Comput.\ Surv.}, vol.~55, no.~10,
pp.~1-36, 2023.

\bibitem{zhang2022fl}
J.~Zhang \emph{et al.}, ``Federated learning for distributed IIoT intrusion
detection using transfer approaches,'' \emph{IEEE Trans.\ Ind.\ Inform.},
vol.~18, no.~12, pp.~8988-8997, 2022.

\bibitem{singh2023}
P.~Singh \emph{et al.}, ``Federated learning based intrusion detection system
for healthcare IoT,'' \emph{IEEE Internet Things J.}, vol.~10, no.~8,
pp.~7254-7264, 2023.

\bibitem{mcmahan2017communication}
B.~McMahan \emph{et al.}, ``Communication-efficient learning of deep networks
from decentralized data,'' in \emph{Proc.\ AISTATS}, vol.~54, pp.~1273-1282,
2017.

\bibitem{li2024fedrl}
W.~Li \emph{et al.}, ``Federated reinforcement learning for adaptive intrusion
response in IoT networks,'' \emph{IEEE Trans.\ Netw.\ Serv.\ Manage.},
vol.~21, no.~2, pp.~1890-1904, 2024.

\bibitem{schulman2017}
J.~Schulman \emph{et al.}, ``Proximal policy optimization algorithms,''
\emph{arXiv:1707.06347}, 2017.

\bibitem{huang2024honeypot}
C.~Huang \emph{et al.}, ``Adaptive honeypot for IoMT: Dynamic deception
with reinforcement learning,'' \emph{IEEE Trans.\ Inf.\ Forensics Security},
vol.~19, pp.~4231-4244, 2024.

\bibitem{qureshi2023dt}
A.~S. Qureshi \emph{et al.}, ``Federated Digital Twin security for smart
hospitals: Anomaly detection without data sharing,'' \emph{IEEE Trans.\ Netw.\
Sci. Eng.}, vol.~10, no.~5, pp.~2893-2905, 2023.

\bibitem{chen2024}
W.~Chen \emph{et al.}, ``Blockchain-enabled digital twin for medical device
authentication in IoMT,'' \emph{IEEE J.\ Biomed.\ Health Inform.}, vol.~28,
no.~1, pp.~245-256, 2024.

\bibitem{li2023}
T.~Li \emph{et al.}, ``Federated optimization in heterogeneous networks,''
\emph{Proc.\ Mach.\ Learn.\ Syst.}, vol.~2, pp.~429-450, 2023.

\bibitem{karimireddy2021scaffold}
S.~P. Karimireddy \emph{et al.}, ``SCAFFOLD: Stochastic controlled averaging
for federated learning,'' in \emph{Proc.\ ICML}, vol.~119, pp.~5132-5143,
2021.

\bibitem{nguyen2022federated}
T.~D. Nguyen \emph{et al.}, ``FLAME: Taming backdoors in federated learning,''
in \emph{Proc.\ USENIX Security}, pp.~1911-1928, 2022.

\bibitem{mirsky2018kitsune}
Y.~Mirsky \emph{et al.}, ``Kitsune: An ensemble of autoencoders for online
network intrusion detection,'' in \emph{Proc.\ NDSS}, 2018.

\bibitem{zhao2023attention}
R.~Zhao \emph{et al.}, ``GRU with temporal attention for low-rate DDoS
detection in SDN environments,'' \emph{IEEE Trans.\ Inf.\ Forensics Security},
vol.~18, pp.~3071-3083, 2023.

\bibitem{ullah2024iot}
I.~Ullah and Q.~H. Mahmoud, ``A two-level flow-based anomaly detection system
for IoT networks using a Siamese neural network,'' \emph{IEEE Access}, vol.~12,
pp.~18221-18234, 2024.

\bibitem{nguyen2023survey}
T.~N. Nguyen, Q.~D. Ngo, and H.~T. Nguyen, ``Graph Neural Network for IoT
intrusion detection: A survey,'' \emph{IEEE Internet Things J.}, vol.~10,
no.~5, pp.~3664-3680, 2023.

\bibitem{genc2022}
Z.~A. Gen{\c{c}}, G.~Lenzini, and S.~E.~J. Ryan, ``Smart honeypots for IoT:
A fuzzy logic approach to adaptive deception,'' in \emph{Proc.\ ARES}, 2022,
pp.~1-10.

\bibitem{kumar2023}
R.~Kumar, S.~S. Gill, and R.~Buyya, ``AI-powered adaptive honeypots for IoT
security: A reinforcement learning approach,'' \emph{Future Gener.\ Comput.\
Syst.}, vol.~138, pp.~131-145, 2023.

\bibitem{nguyen2021}
K.~K. Nguyen and V.~J. Reddi, ``Deep reinforcement learning for cyber security:
A survey,'' \emph{ACM Comput.\ Surv.}, vol.~54, no.~1, pp.~1-36, 2021.

\bibitem{zhou2023}
Y.~Zhou \emph{et al.}, ``DeepFort: Reinforcement learning based DDoS mitigation
for IoT networks,'' \emph{IEEE Trans.\ Netw.\ Serv.\ Manage.}, vol.~20, no.~2,
pp.~1234-1247, 2023.

\bibitem{salim2022}
M.~M. Salim \emph{et al.}, ``A blockchain-enabled secure digital twin framework
for early botnet detection in IIoT environment,'' \emph{Sensors}, vol.~22,
no.~16, p.~5985, 2022.

\bibitem{khan2022a}
I.~A. Khan \emph{et al.}, ``Fed-Inforce-Fusion: A federated reinforcement-based
fusion model for security and privacy protection of IoMT networks against
cyber-attacks,'' \emph{Inf.\ Fusion}, vol.~90, pp.~438-452, 2023.

\bibitem{cicddos2019}
I.~Sharafaldin \emph{et al.}, ``Developing realistic distributed denial of
service (DDoS) attack dataset and taxonomy,'' in \emph{Proc.\ ICCST}, 2019,
pp.~1-8.

\bibitem{toniot2020}
A.~Alsaedi \emph{et al.}, ``TON\_IoT Telemetry Dataset: A new generation
dataset of IoT and IIoT for data-driven intrusion detection systems,''
\emph{IEEE Access}, vol.~8, pp.~165130-165150, 2020.

\bibitem{popoola2021}
S.~I. Popoola \emph{et al.}, ``Federated deep learning for zero-day botnet
attack detection in IoT-Edge devices,'' \emph{IEEE Internet Things J.}, vol.~9,
no.~5, pp.~3930-3944, 2022.

\bibitem{naeem2023}
F.~Naeem, M.~Ali, and G.~Kaddoum, ``Federated-learning-empowered
semi-supervised active learning framework for intrusion detection in ZSM,''
\emph{IEEE Commun.\ Mag.}, vol.~61, no.~2, pp.~88-94, 2023.

\bibitem{li2023ids}
J.~Li \emph{et al.}, ``An efficient federated learning system for network
intrusion detection,'' \emph{IEEE Syst.\ J.}, vol.~17, no.~3,
pp.~4819-4829, 2023.

\bibitem{chen2020}
Y.~Chen \emph{et al.}, ``FedHealth: A federated transfer learning framework
for wearable healthcare,'' \emph{IEEE Intell.\ Syst.}, vol.~35, no.~4,
pp.~83-93, 2020.

\bibitem{li2020}
B.~Li \emph{et al.}, ``DeepFed: Federated deep learning for intrusion detection
in industrial cyber-physical systems,'' \emph{IEEE Trans.\ Ind.\ Inform.},
vol.~17, no.~8, pp.~5615-5624, 2021.

\bibitem{schneble2019}
W.~Schneble and G.~Thamilarasu, ``Attack detection using federated learning in
medical cyber-physical systems,'' in \emph{Proc.\ ICCCN}, 2019, pp.~1-8.

\bibitem{nguyen2019}
T.~D. Nguyen \emph{et al.}, ``D{\`\i}ot: A federated self-learning anomaly
detection system for IoT,'' in \emph{Proc.\ IEEE ICDCS}, 2019, pp.~756-767.

\bibitem{hossen2022}
M.~N. Hossen \emph{et al.}, ``Federated machine learning for detection of skin
diseases and enhancement of internet of medical things (IoMT) security,''
\emph{IEEE J.\ Biomed.\ Health Inform.}, vol.~26, no.~10,
pp.~4982-4991, 2022.

\bibitem{xu2020}
Y.~Xu \emph{et al.}, ``Feature data processing: Making medical data fit deep
neural networks,'' \emph{Future Gener.\ Comput.\ Syst.}, vol.~109,
pp.~149-157, 2020.

\bibitem{fan2022}
J.~Fan \emph{et al.}, ``Federated learning driven secure internet of medical
things,'' \emph{IEEE Wireless Commun.}, vol.~29, no.~2, pp.~68-75, 2022.

\bibitem{dwork2014dp}
C.~Dwork and A.~Roth, ``The algorithmic foundations of differential privacy,''
\emph{Found.\ Trends Theor.\ Comput.\ Sci.}, vol.~9, no.~3-4,
pp.~211-407, 2014.

\bibitem{li2020fedprox}
T.~Li \emph{et al.}, ``Federated optimization in heterogeneous networks,''
in \emph{Proc.\ MLSys}, vol.~2, pp.~429-450, 2020.

\bibitem{lundberg2017shap}
S.~M. Lundberg and S.-I. Lee, ``A unified approach to interpreting model
predictions,'' in \emph{Proc.\ NeurIPS}, vol.~30, pp.~4765-4774, 2017.

\bibitem{ribeiro2016lime}
M.~T. Ribeiro, S.~Singh, and C.~Guestrin, ```Why should I trust you?':
Explaining the predictions of any classifier,'' in \emph{Proc.\ KDD},
pp.~1135-1144, 2016.

\bibitem{selvaraju2020gradcam}
R.~R. Selvaraju \emph{et al.}, ``Grad-CAM: Visual explanations from deep
networks via gradient-based localization,'' \emph{Int.\ J.\ Comput.\ Vis.},
vol.~128, no.~2, pp.~336-359, 2020.

\bibitem{wachter2017counterfactual}
S.~Wachter, B.~Mittelstadt, and C.~Russell, ``Counterfactual explanations
without opening the black box: Automated decisions and the GDPR,''
\emph{Harvard J.\ Law Technology}, vol.~31, no.~2, pp.~841-887, 2018.

\end{thebibliography}
\end{document}